\pdfoutput=1
\documentclass[11pt]{article}

\usepackage[preprint]{acl}

\usepackage{times}
\usepackage{latexsym}

\usepackage[T1]{fontenc}
\usepackage[utf8]{inputenc}

\usepackage{microtype}

\usepackage{inconsolata}

\usepackage{graphicx}

\usepackage{colortbl} % 引入colortbl包

\usepackage{multirow}
\usepackage{multicol}
\usepackage{booktabs}
\usepackage{graphicx}
\usepackage{tikz}
\usepackage{paralist}
\usepackage{booktabs}
\usepackage{amsfonts}
\usepackage{CJKutf8}
\usepackage{subfigure}

\usepackage[ruled,vlined]{algorithm2e}
\usepackage{amssymb}
\usepackage{enumitem}
\usepackage{setspace}%行间距
\usepackage{amsmath} 
\usepackage{amssymb}
\usepackage{xcolor}
\usepackage{graphicx}
\usepackage{url}
\usepackage{booktabs}
\usepackage{amssymb}
\usepackage{bbding}
\usepackage{pifont}
\usepackage{wasysym}
\usepackage{utfsym}
\usepackage{fontawesome}
\usepackage{footnote} 
\usepackage{makecell}
\usepackage{wrapfig}
\usepackage{ulem}

\usepackage{algpseudocode}

\title{S$^{3}$-CoT: Self-Sampled Succinct Reasoning Enables \\ Efficient Chain-of-Thought LLMs}

\author{\textbf{
Yanrui Du\textsuperscript{1},
Sendong Zhao\textsuperscript{1}\thanks{\llap{}Corresponding Author.}, 
Yibo Gao\textsuperscript{1}, 
Danyang Zhao\textsuperscript{1},
Qika Lin\textsuperscript{1},
Ming Ma\textsuperscript{1},
} \\
\textbf{
Jiayun Li\textsuperscript{1},
Yi Jiang\textsuperscript{1},
Kai He\textsuperscript{2},
Qianyi Xu\textsuperscript{2},
Bing Qin\textsuperscript{1},
Mengling Feng\textsuperscript{2}
}\\
    \textsuperscript{1}Harbin Institute of Technology, Harbin, China \\  
    \textsuperscript{2}National University of Singapore, Singapore\\
     \{yrdu,sdzhao\}@ir.hit.edu.cn\\
}

\begin{document}

\maketitle

\begin{abstract}

Large language models (LLMs) equipped with chain-of-thought (CoT) achieve strong performance and offer a window into LLM behavior. 
However, recent evidence suggests that improvements in CoT capabilities often come with redundant reasoning processes, motivating a key question: can LLMs acquire a ``fast-thinking'' mode analogous to human System 1 reasoning?
To explore this, our study presents a self-sampling framework based on activation steering for efficient CoT learning. 
Our method can induce style-aligned and variable-length reasoning traces from target LLMs themselves without any teacher guidance, thereby alleviating a central bottleneck of SFT-based methods—the scarcity of high-quality supervision data.
Using filtered data by gold answers, we perform SFT for efficient CoT learning with (i) a human-like dual-cognitive system, and (ii) a progressive compression curriculum.
Furthermore, we explore a self-evolution regime in which SFT is driven solely by prediction-consistent data of variable-length variants, eliminating the need for gold answers.
Extensive experiments on math benchmarks, together with cross-domain generalization tests in medicine, show that our method yields stable improvements for both general and R1-style LLMs.
Our data and model checkpoints can be found at \url{https://github.com/DYR1/S3-CoT}.

\end{abstract}

% \textbf{Warning: This paper presents malicious examples that may be offensive and upsetting.}

\newcommand{\Tabi}[2]{\begin{tabular}{@{}#1@{}}#2\end{tabular}}

% \vspace{-0.1cm}
\section{Introduction}\label{sec_intro}
% \vspace{-0.1cm}

Chain-of-thought (CoT) has become a standard mechanism for eliciting multi-step reasoning in large language models (LLMs), substantially improving performance on complex tasks \citep{wei2022chain,yao2023tree,besta2024graph, wang2022self}. 
More recently, the field has shifted toward internalizing such reasoning behaviors into LLMs themselves via post-training pipelines, aiming to make strong reasoning the default rather than prompt-contingent \citep{zhao2024marco,jaech2024openai,guo2025deepseek,yu2024distilling}. 
However, once reasoning is internalized, the generated reasoning traces often become overly long and redundant, inflating latency and cost even on easy instances \citep{wu2025more,liu2025laser}. 
This motivates methods that \textit{compress reasoning traces while preserving reasoning ability}.

% Subsequent inference-time strategies further increase effective test-time compute by aggregating multiple trajectories \citep{yao2023tree,besta2024graph, wang2022self}. 

% , for example, self-consistency improves reliability
 % by externalizing intermediate steps 

To achieve this, existing work can be grouped into three categories.
(1) \textbf{Prompt-based control} constrains reasoning length via explicit budgets or specialized templates \citep{han2025token,nayab2024concise}.
While lightweight, these methods are highly sensitive to prompt wording, and often require task- or model-specific tuning.
(2) \textbf{SFT-based methods} fine-tune LLMs with curated concise traces as supervision \citep{munkhbat2025self}.
Their primary bottleneck is \emph{supervision data}: collecting high-quality and variable-length CoTs is expensive and difficult.
Both C3oT \citep{kang2025c3ot} and CoT-Valve \citep{ma2025cot} require guidance from external tools or teacher LLMs to achieve this.
Such teacher-dependent pipelines can be brittle, as CoT verbosity and stylistic conventions vary widely across LLM families.
(3) \textbf{RL-based methods} explicitly optimize the length–accuracy trade-off by shaping rewards or enforcing token constraints during training \citep{hou2025thinkprune,liu2025laser,tu2025learning}.
Although effective, RL typically incurs a higher computational cost and is sensitive to reward design and training stability.

% heuristically steering LLMs toward shorter traces 

\begin{figure*}[t]
\centering
\includegraphics[scale=0.6]{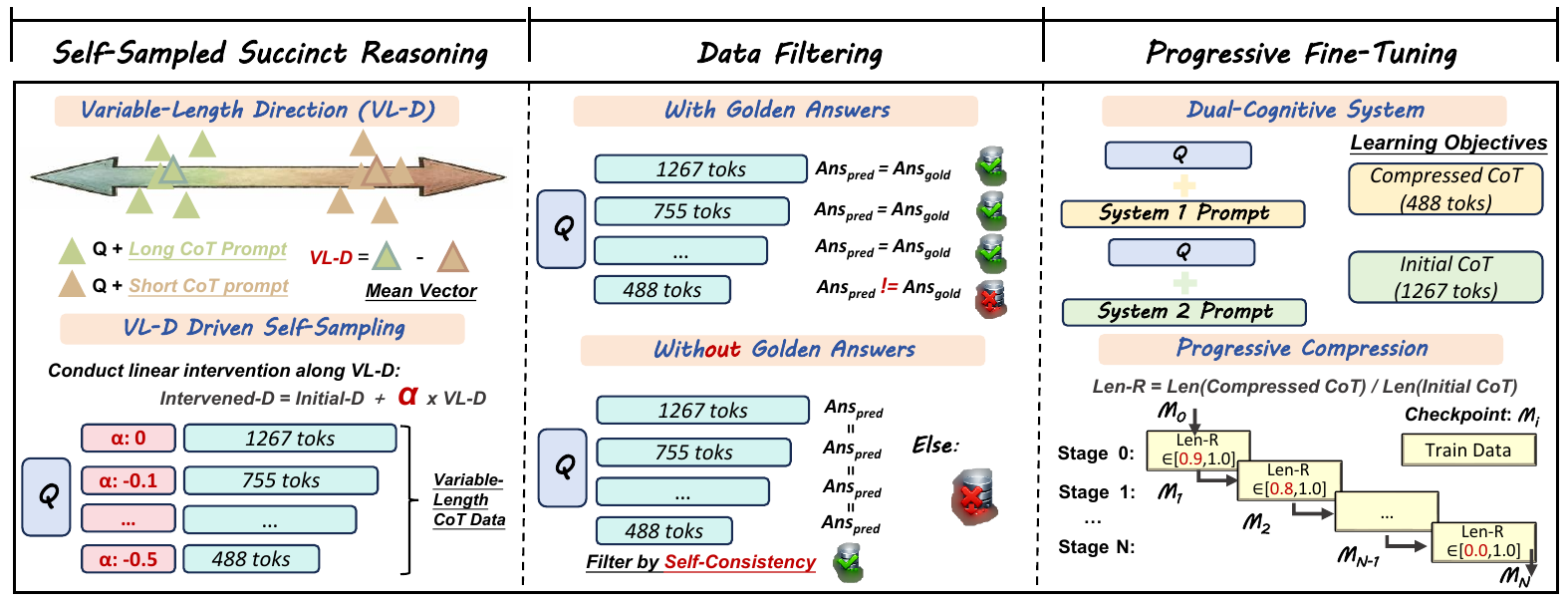}
\vspace{-0.2cm}
\caption{A self-sampling framework for efficient CoT learning. Our study (1) samples variable-length CoT data via intervention along VL-D; (2) filters data via answer or self-consistency verification; and (3) achieves efficient CoT internalization via a dual-cognitive system and progressive compression curriculum.}
\vspace{-0.5cm}
\label{fig_overall_framework}
\end{figure*}

% For instance, C3oT\citep{kang2025c3ot} relies on external tools (e.g., GPT-4) to prune redundant steps, and CoT-Valve\citep{ma2025cot} trains a controllable LoRA module on QwQ-32B to distill variable-length CoT data.

In our study, we target \textbf{the gap left by SFT-based methods}: how to obtain high-quality, style-aligned, variable-length CoT data without any teacher guidance.
Inspired by activation steering, we propose S$^{3}$-CoT, a self-sampling framework for efficient CoT learning. 
Fig.~\ref{fig_overall_framework} presents the key idea of our framework.
Activation steering posits that LLM behaviors can be modulated via interventions along (approximately) linear directions in representation space \citep{tigges2023linear,zou2023representation,rimsky2024steering,turner2023steering}.
Building on these insights, we conduct targeted analyses to identify a variable-length direction (VL-D) that governs CoT lengths.
Guided by the intervention settings revealed in our probe analysis, we sample variable-length CoT traces directly from target LLMs themselves.
To further ensure data quality, we apply either answer or self-consistency verification \citep{wang2022self}.
Notably, for the latter, we retain prediction-consistent data of variable-length CoT variants, yielding a fully self-evolved data acquisition process. 
Our analysis shows that samples retained via self-consistency typically achieve near-perfect accuracy.
During SFT, we adopt a dual-cognitive system and a progressively compressed curriculum, enabling LLMs to acquire fast-thinking capabilities while avoiding performance degradation caused by over-compression.

% This design helps LLMs equip themselves with fast-thinking while avoiding performance degradation from over-compression.

In our experiments, extensive evaluation on math and medical benchmarks shows that our method consistently outperforms prompt-control and SFT-based baselines and achieves performance competitive with RL-based baselines.
Notably, our method exhibits strong adaptability across various LLMs (general and R1-style LLMs\footnote{In our study,  we term LLMs that emit ``<think></think>'' reasoning traces as R1-style LLMs, and LLMs with standard outputs as general LLMs.}) and datasets, a setting that has rarely been validated in prior work.
Overall, our contributions can be summarized as follows:
1) We propose S$^{3}$-CoT to alleviate the data-scarcity bottleneck of SFT-based methods, via a standardized pipeline that samples high-quality, variable-length CoTs from target LLMs themselves.
2) Leveraging self-sampled data, we enable efficient CoT internalization through SFT, providing an early exploration of LLM self-evolution.
3) Extensive experiments show our method achieves superior or competitive performance, with strong adaptability across various LLMs and datasets.

\vspace{-0.1cm}
\section{Related Work}\label{sec_related}
\vspace{-0.1cm}

\subsection{Efficient CoT Internalization }
\vspace{-0.1cm}

Existing efforts largely fall into three paradigms: prompt control, SFT-based, and RL-based optimization. 
Prompt control imposes inference-time constraints by injecting explicit length cues or enforcing structured reasoning formats, offering a lightweight way to shorten CoT \citep{nayab2024concise,renze2024benefits,xu2025chain,han2025token}. 
SFT-based methods aim to internalize concise reasoning by fine-tuning LLMs on succinct CoT data, enabling shorter CoT without relying on prompts \citep{liu2024can, yu2024distilling,kang2025c3ot,ma2025cot,munkhbat2025self,xia2025tokenskip}.
RL-based methods further treat conciseness as an optimization objective, explicitly balancing accuracy and length through reward design, and have shown strong effectiveness \citep{hou2025thinkprune,liu2025laser,tu2025learning,yi2025shorterbetter,cheng2025optimizing,luo2025o1,arora2025training}. 
Such RL pipelines are most commonly applied to R1-style LLMs in existing work, whereas their applicability and stability for general LLMs are less explored.
A more detailed description of existing methods can be found in our Appendix ~\ref{app_related}.

\vspace{-0.1cm}
\subsection{Activation Steering}
\vspace{-0.1cm}
Activation steering \citep{zou2023representation} aims to control LLM behavior via intervention along approximately linear directions in representation space: 
Activation Addition demonstrates that adding a direction vector can induce target attributes \citep{turner2023steering}, and Contrastive Activation Addition constructs steering vectors from contrastive residual differences to modulate behaviors like sycophancy \citep{rimsky2024steering}.
Related work further develops concept-direction discovery (e.g., sentiment directions \citep{tigges2023linear} and refusal directions \citep{arditi2024refusal}), and inference-time interventions that change LLM outputs by targeted activation edits \citep{li2023inference, azizi2025activation,tang2025unlocking,du2025anchoring}.

% In our study, we leverage activation steering as a data-sampling tool to facilitate efficient CoT learning. 
% This line of exploration suggests an LLM-level capacity for self-evolution, and to the best of our knowledge, we are among the earliest teams to systematically investigate this pathway.

% Our study reveals the existence of variable-length direction and leverages this property to sample variable-length reasoning data for training efficient CoT LLMs.

% \clearpage

\vspace{-0.1cm}
\section{Self-Sampled Succinct Reasoning}\label{sec_sample}
\vspace{-0.1cm}

\begin{figure*}[ht]
\centering
\subfigure[Analysis on Qwen2.5$_{7B}$.]
{\includegraphics[scale=0.65]{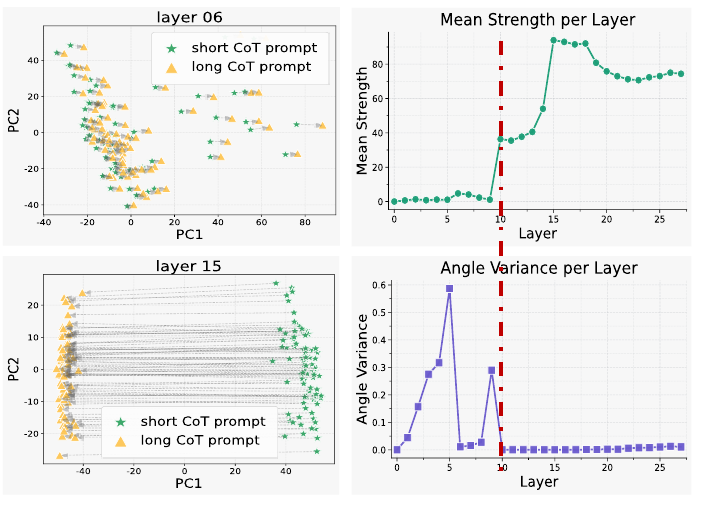}
}
% \hfill
\subfigure[Analysis on Deepseek-R1$_{7B}$.]
{\includegraphics[scale=0.65]{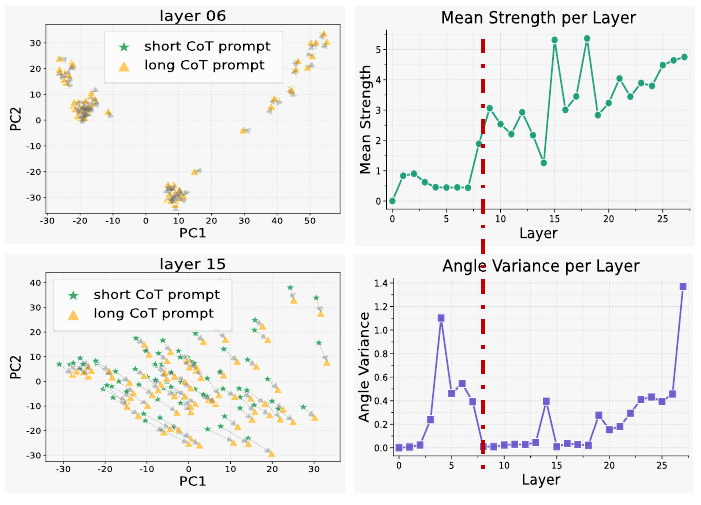}
}
\vspace{-0.2cm}
\caption{Analysis of VL-D properties. We provide PCA-based visualizations and quantify how the mean separation strength and angle variance metric vary across layers.
Visualizations across all layers under various LLMs are in Fig.~\ref{app_vis_qw2.5},~\ref{app_vis_dsqw},~\ref{app_vis_ll3}, and ~\ref{app_vis_qw3}, respectively.
Analysis on LLaMA3$_{8B}$ and Qwen3-Think$_{4B}$ are in Fig.~\ref{app_fig_ana}
.}
\label{fig_ana}
\vspace{-0.4cm}
\end{figure*}

In this section, we primarily answer three questions: 
\textbf{Q1:} Does a length-controlled linear direction exist in LLMs' representation space? 
\textbf{Q2:} How can we sample our expected data via intervention along this direction? 
\textbf{Q3:} Can this method facilitate the sampling of high-quality, variable-length CoT data?
Our experiments are conducted primarily on GSM8K \citep{cobbe2021training} and span both general LLMs (Qwen2.5$_{7B}$ \cite{qwen2.5} /LLaMA3$_{8B}$ \cite{dubey2024llama}) and R1-style LLMs (DeepSeek-R1$_{7B}$ \citep{deepseekai2025deepseekr1incentivizingreasoningcapability}/Qwen3-Think$_{4B}$ \citep{qwen3technicalreport}).

\vspace{-0.2cm}
\subsection{Identification of VL-D}
\paragraph{VL-D Extraction.} 
Prior direction-extraction methods typically derive representations from contrastive instruction pairs about a single attribute \citep{zou2023representation,arditi2024refusal,tigges2023linear}.
Similarly, for length attribution, we append long and short CoT prompts to each instruction ($x \in D$), resulting in two sets $D_{L}$ and $D_{S}$ ($(x_{l},x_{s}) \in (D_{L},D_{S})$).
To formalize the direction extraction process, we begin with the decoder-only transformer architecture.
Each input sequence $x = (x_1, x_2, \ldots, x_n) \in \mathcal{V}^n$ is mapped to output probabilities ($y \in \mathbb{R}^{n \times |\mathcal{V}|}$).
The residual stream activation of token $i$ at the start of layer $l$ is denoted as $\mathbf{h}^{(l)}_i \in \mathbb{R}^{d_{\text{model}}}$, initialized with its embedding $\mathbf{h}^{(1)}_i = \text{Embed}(x_i)$. 
Each layer applies both attention and MLP transformations:
$
\tilde{\mathbf{h}}^{(l)}_i = \mathbf{h}^{(l)}_i + \text{Attn}^{(l)}(\mathbf{h}^{(l)}_{1:n}),
$
$
\quad \mathbf{h}^{(l+1)}_i = \tilde{\mathbf{h}}^{(l)}_i + \text{MLP}^{(l)}(\tilde{\mathbf{h}}^{(l)}_i).
$
The variable-length direction can be extracted using the difference-in-means method \citep{marks2023geometry,panickssery2023steering}. 
For each layer $l \in [L]$ and final token position $n$, activations $(\mathbf{h}^{(l)}_n(x_{l}) , \mathbf{h}^{(l)}_n(x_{s}))$ over $(x_{l},x_{s}) \in (D_{L},D_{S})$ are obtained, and the corresponding difference-in-means vector can be calculated as:
$
\mathbf{u}^{(l)} = \mathbf{h}^{(l)}_n(x_{l}) - \mathbf{h}^{(l)}_n(x_{s}),
$
$
d^{(l)} = \mathbb{E}_{u\sim U^{(l)}}[u].
$

% $
% \quad d^{(l)} = \frac{1}{|U^{(l)}|} \sum_{u^{(l)}_{j} \in U^{(l)}} u^{(l)}_{j}.
% $

\paragraph{Visualization Analysis.}
To further investigate the nature of VL-D, we apply PCA \citep{hotelling1933analysis} for dimensionality reduction and plot the extracted direction among each pair $(x_{l},x_{s})$. 
Here, we sample 100 data points from GSM8K, retaining only those where our appended CoT prompt significantly influences the length.
Such a limitation allows us to filter out the impact of those cases where LLMs fail to follow instructions.
As shown in Fig.~\ref{fig_ana}, we present visualizations for the 6th and 15th layer under two LLMs, with visualizations for other LLMs and all layers provided in Fig.~\ref{app_fig_ana}, Fig.~\ref{app_vis_qw2.5},~\ref{app_vis_dsqw},~\ref{app_vis_ll3}, and ~\ref{app_vis_qw3}. 
Two key phenomena emerge:
\begin{itemize}[leftmargin=*,noitemsep,topsep=0pt]
\item 
\textbf{Layer-wise Separation Emergence:} Starting from the middle layers, a clear separation between $x_{l}$ and $x_{s}$ can be observed, while such separation is less pronounced in earlier layers. This suggests that the length-controlled directions may begin to emerge in the middle layers.
\item 
\textbf{Parallelism of Directions:}
Once the separation emerges, the directions extracted for each sample pair are highly parallel. 
This indicates the presence of a length-controlled direction, independent of the individual sample.
\end{itemize} 
% These observations answer Q1: there exists a length-control linear direction in LLMs' representation space.

% Given that LLMs exhibit different capabilities in following instructions, they will not always adhere to the length constraint prompt. 

\vspace{-0.1cm}
\subsection{Intervention along VL-D}

\paragraph{Quantitative Metrics.}

Although visualization analyses suggest the existence of variable-length directions, fine-grained intervention requires quantitative metrics to better understand their properties.
Therefore, we introduce mean separation strength and angle variance metrics to monitor PCA-reduced features. 
The mean separation strength metric computes the L2 distance in each pair, and for the $l^{th}$ layer, it can be calculated as:
\begin{equation}
\small
\mathbf{S^{(l)}} = \frac{1}{|D_{L}|} \sum_{(x_{l},x_{s}) \in (D_{L},D_{S})} \| \mathbf{h}^{(l)}_{pca}(x_{l}) -  \mathbf{h}^{(l)}_{pca}(x_{s}) \|_2.
\end{equation}
Meanwhile, the angle variance metric calculates the angle variance of each sample pair's direction relative to their mean direction.
For the $l^{th}$ layer, by normalization, we can obtain unit vectors $\mathbf{\bar{u}}^{pca}_{i}$ for each pair and their mean $ \mathbf{v}^{pca}$.
The cosine value between each unit vector and the mean is: $\cos \theta_i = \mathbf{\bar{u}}^{pca}_{i} \cdot \mathbf{v}^{pca}$.
The angle \(\theta_i \) can be calculated by the inverse cosine function: $\theta_i = \arccos(\cos \theta_i)$ and the angle variance can be further calculated as:
\begin{equation}
\small
\sigma_{\theta}^2 = \frac{1}{m-1} \sum_{i=1}^{m} \left( \theta_i - \bar{\theta} \right)^2.
\end{equation}
where $\bar{\theta}$ represents the average value of $\theta$.
The line chart of Fig.~\ref{fig_ana} and Fig.~\ref{app_fig_ana} (in Appendix) presents the change of mean separation strength and angle variance across all layers.
We observe that starting from certain middle layers (marked by the red dashed lines), there is a significant separation in each pair, and the directions among pairs become highly parallel.
For Qwen2.5$_{7B}$, LLaMA3$_{8B}$, and Qwen3-Think$_{4B}$, this phenomenon persists until the final layer, while for Deepseek-R1$_{7B}$, it remains relatively stable in the middle layers.
We guess that the extensive incremental training of DeepSeek-R1$_{7B}$ may have introduced instability in its internal properties.
Nevertheless, across various LLMs, we observe that there exists a length-controlled linear direction starting from the middle layers.

% this is due to Deepseek-R1$_{7B}$ undergoing extensive incremental training, which may have disrupted some of its inherent properties.
% Overall, across various LLMs, there exists a length-controlled linear direction starting from the middle layers.

% Overall, we can conclude that across various LLMs, there exists a linear direction controlling the CoT length starting from the middle layers.

\begin{figure*}[ht]
\centering
\subfigure[Analysis on Qwen2.5$_{7B}$.]
{\includegraphics[scale=0.52]{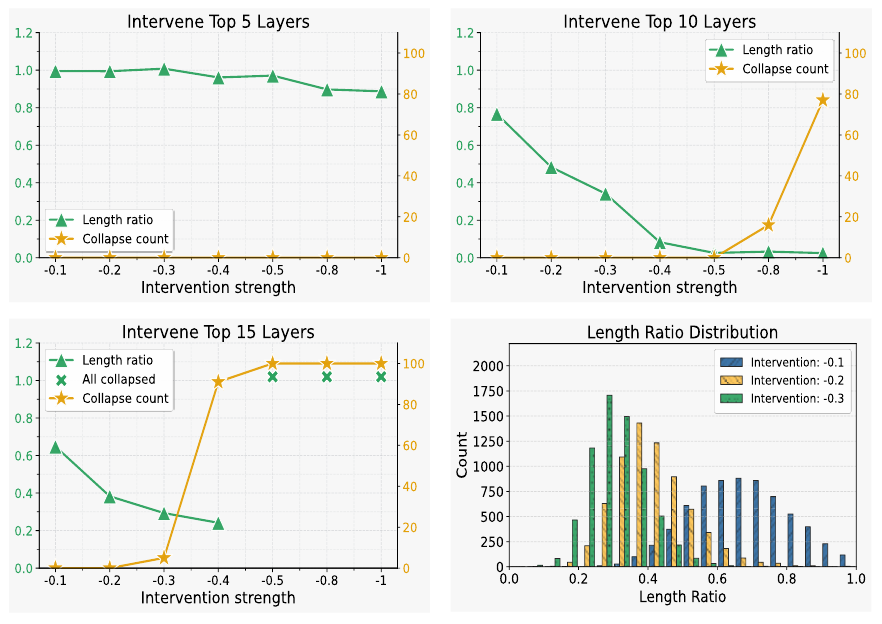}
}
% \hfill
\subfigure[Analysis on Deepseek-R1$_{7B}$.]
{\includegraphics[scale=0.52]{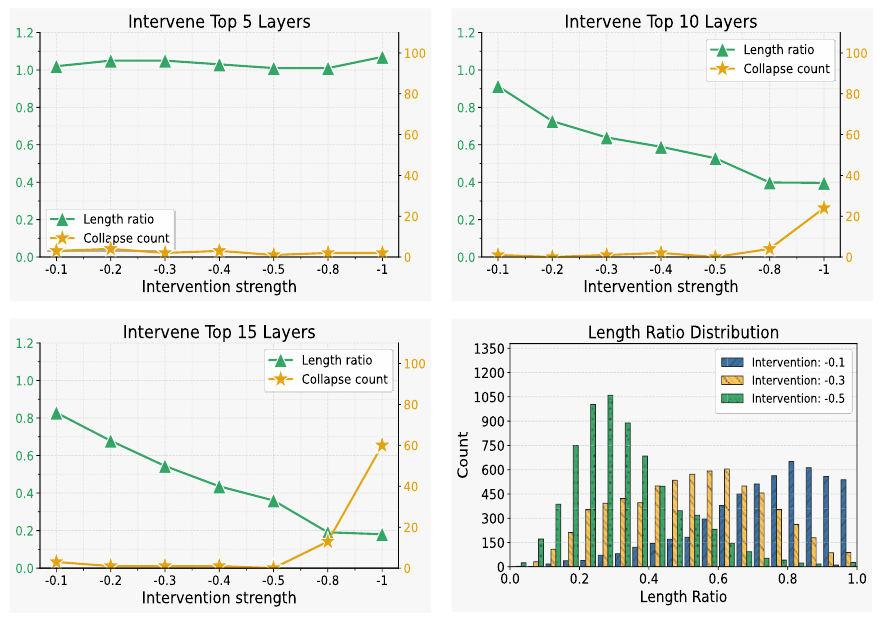}
}
\vspace{-0.4cm}
\caption{Probe experiments on intervention layers and strength. Green: average Len-R; Yellow: number of collapsed samples; Green ``×'': all samples collapse. Bottom-right: Len-R distribution under large-scale sampling. Results for LLaMA3$_{8B}$ and Qwen3-Think$_{4B}$ are in Fig.~\ref{app_fig_ratio}, and results for other intervention settings are in Fig.~\ref{app_interve_other_sets}.}
\label{fig_ratio}
\vspace{-0.4cm}
\end{figure*}

\paragraph{Probing Analysis.}
Our goal is to induce variable-length CoT data via interventions along VL-D. Following prior work \citep{arditi2024refusal}, the intervention can be modeled as a linear operation. 
Given an input $x$, we modify the hidden state at $i^{th}$ token in $l^{th}$ layer as:
$
\mathbf{h}^{(l)}_{i}(x) \leftarrow \mathbf{h}^{(l)}_{i}(x) + \alpha \times \mathbf{d}^{(l)},
$
where $\alpha$ is a tunable intervention strength.
Furthermore, we conduct a probing analysis to guide the choice of (i) intervention layers and (ii) intervention strength.
Specifically, we define \emph{Length-Ratio (Len-R) as the ratio between the post-intervention output length and the initial length}, which reflects the effectiveness of interventions.
And we define the layer at which the VL-D first emerges (marked by red dashed lines in Fig.~\ref{fig_ana} and ~\ref{app_fig_ana}) as the anchor layer.
For layer selection, we intervene on a contiguous block of layers starting from the anchor layer.
Taking Qwen2.5$_{7B}$ as an example, the anchor layer is the 10$^{th}$ layer, ``Top 1'' corresponds to intervening on layers [10,11), and ``Top 5'' corresponds to [10, 15) layers, and so forth.
For intervention strength, since our goal is to obtain shorter CoTs for efficient learning, we consider $\alpha \in \{-0.1, -0.2, -0.3, -0.4, -0.5, -0.8, -1\}$.

% , and evaluate an additional 100 samples

Our analysis evaluates an additional 100 samples, and Fig.~\ref{fig_ratio} and Fig.~\ref{app_fig_ratio} (in Appendix) summarize the probing results.
We observe that: (1) weak interventions—either intervening on few layers or a small $\alpha$ —may fail to shorten CoTs, with Len-R remaining close to 1;
(2) overly strong interventions—either intervening on many layers or a large $\alpha$—may trigger generation collapse, resulting in repetitive outputs (marked by green ``x'').
These results highlight the importance of careful hyperparameter selection.
Empirically, for general LLMs, intervening on top 5–10 layers with $|\alpha| \leq 0.5$ is typically stable, whereas for R1-style LLMs, intervening on top 15 layers with $|\alpha| \leq 0.5$ yields stable performance.
While these trends provide coarse guidance, we do not observe a universally optimal setting across LLMs. 
Therefore, we advocate a standardized probing step on a small pilot set prior to large-scale intervention.
According to probing results, to sample variable-length CoT data, our study intervenes on top 10/5/15/15 layers for Qwen2.5$_{7B}$/LLaMA3$_{8B}$/DeepSeek-R1$_{7B}$/Qwen3-Think$_{4B}$ with $\alpha \in$ (\{-0.1,-0.2,-0.3\}), (\{-0.1,-0.3,-0.5\}), (\{-0.1,-0.3,-0.5\}), and (\{-0.1,-0.3,-0.5\}), respectively.

% These probing results help us answer Q2: we have established a standardized procedure to guide interventions along VL-D.

% Requires: \usepackage{booktabs}

\vspace{-0.1cm}
\subsection{Verification of Data Quality}

Since our sampled data are induced from target LLMs themselves, they can naturally keep style-aligned output. 
Accordingly, we focus on two other aspects of data quality: correctness and variable-length behavior.

\paragraph{For correctness.} 
When gold answers are available, we adopt an answer verification scheme to filter data, retaining only those whose predictions match gold answers.
However, in some practical settings, annotating answers is expensive.
Therefore, as shown in Fig.~\ref{fig_overall_framework}, we explore a self-consistency verification scheme: only samples whose predictions remain consistent across variable-length variants are retained.
Tab.~\ref{tab:self-consistency} reports the number and the accuracy of GSM8K samples retained under this scheme.
Interestingly, across various LLMs, retained samples typically achieve near-perfect accuracy.
But the limitation of this scheme is that sampling efficiency will be affected by the underlying LLM capability.
For example, for LLaMA3$_{8B}$, only 517 out of 6,838 samples are retained, whereas other stronger LLMs exhibit substantially higher sampling efficiency.

\paragraph{For variable-length behavior.} The bottom-right corner of Fig.~\ref{fig_ratio} and Fig.~\ref{app_fig_ratio} (in Appendix) present the Len–R distribution of our sampled data. 
As the intervention strength $|\alpha|$ increases, the overall distribution shifts left, indicating shorter CoT on average. 
This trend confirms that our method can effectively sample variable-length CoT data.

% These discussions help us answer Q3: our proposed method can sample high-quality, variable-length CoT data.
% But this scheme may reduce sampling efficiency, depending on the underlying LLM capability.

\begin{table}[t]
\centering
\small
\setlength{\tabcolsep}{4pt}
\begin{tabular}{lcccc}
\toprule
LLM & \#Total & \#Retained & \#Correct & Acc.\\
\midrule
DeepSeek-R1$_{7B}$   & \multirow{4}{*}{6,838} & 5,655 & 5,648 & 99.88\% \\
Qwen3-Think$_{4B}$    &  & 6,468 & 6,449 & 99.71\% \\
Qwen2.5$_{7B}$  &  & 4,564 & 4,560 & 99.91\% \\
LLaMA3$_{8B}$    &  &   517 &   516 & 99.81\% \\
\bottomrule
\end{tabular}
\caption{The number and accuracy of samples retained by self-consistency verification under various LLMs.}
\vspace{-0.4cm}
\label{tab:self-consistency}
\end{table}

Overall, this section identifies the variable-length direction, conducts probe analysis on intervention settings, and validates the high quality of our sampled data. 
% Next, we will answer the question of whether our sampled data can enable efficient CoT LLMs.

% investigate whether the data collected by our method can be used to train CoT-enabled LLMs efficiently.

% To ensure correctness, we introduce answer verification and a self-consistency verification mechanism. As illustrated in the middle of Figure 1, when ground-truth answers are available, we retain only those sampled instances whose predictions match the reference answers.

% However, in many practical settings, annotating answers is expensive. We therefore adopt a self-consistency verification mechanism: we keep only samples whose predictions remain consistent across variable-length variants. Table 1 reports the accuracy of samples selected under this mechanism. We observe that, across different models, the selected samples typically achieve very high accuracy—on GSM8K, the accuracy can reach nearly 100%. Meanwhile, we also find that self-consistency filtering may reduce sampling efficiency, depending on the underlying model capability. For example, for Llama 3, only 517 out of 6,838 samples are retained, whereas other models generally exhibit substantially higher sampling efficiency.

\vspace{-0.2cm}
\section{Efficient CoT LLMs}\label{sec_main_exp}
\vspace{-0.1cm}

In this section, we answer the question of whether self-sampled data can enable efficient CoT LLMs.

\begin{table*}[t]
\centering
\small
\setlength{\tabcolsep}{2.0pt}
\begin{tabular}{lccccc|ccccc|c}
\toprule
\multirow{2}{*}{Method} &
\multicolumn{5}{c|}{Accuracy$\uparrow$} &
\multicolumn{5}{c|}{Length$\downarrow$} &
\multirow{2}{*}{AES$\uparrow$} \\
\cmidrule(lr){2-6}\cmidrule(lr){7-11}
& GSM8K & MATH & AMC23 & AIME24 & AVG. & GSM8K & MATH & AMC23 & AIME24 & AVG. & \\
\midrule
\multicolumn{12}{c}{\textbf{Qwen2.5$_{7B}$}} \\
% \midrule
Standard$_p$        & 93.33\% & 72.67\% & 43.33\% & 7.78\%  & 54.28\% & 289.82 & 559.49 & 846.49 & 996.75  & 673.14  & -- \\
Efficient$_p$       & 89.67\% & 70.00\% & 44.17\% & 7.78\%  & 52.90\% & 107.15 & 300.83 & 573.44 & 741.21  & 430.66  & 0.11 \\
\rowcolor{blue!5}
TokenSKIP         & 90.83\% & 74.67\% & 45.00\% & 2.33\%  & 53.21\% & 260.63 & 512.12 & 766.95 & 842.71  & 595.60  & -0.08 \\
\rowcolor{blue!5}
CoT-Valve          & 90.50\% & 71.00\% & 37.50\% & 6.68\%  & 51.42\% & 298.82 & 619.19 & 900.23 & 1068.58 & 721.71  & -0.60 \\
\rowcolor{blue!5}
C3oT                 & 93.50\% & 71.33\% & 51.67\% & 5.56\%  & 55.52\% & 291.22 & 536.77 & 788.60 & 866.80  & 620.85  & 0.19 \\
\rowcolor{blue!5}
S$^{3}$-CoT         & 93.17\% & 70.50\% & 45.83\% & 12.22\% & 55.43\% & 182.80 & 426.91 & 678.80 & 800.62  & 522.29  & 0.33 \\
\rowcolor{blue!5}
S$^{3}$-CoT$^{sc}$  & 92.50\% & 69.67\% & 46.67\% & 11.11\% & 54.99\% & 184.10 & 433.43 & 687.33 & 831.47  & 534.08  & 0.27 \\
\midrule
\multicolumn{12}{c}{\textbf{DeepSeek-R1$_{7B}$}} \\
% \midrule
Standard$_p$        & 93.33\% & 92.33\% & 90.83\% & 51.11\% & 81.90\% & 1710.27 & 4261.18 & 6224.23 & 14061.22 & 6564.23 & -- \\
Efficient$_p$        & 84.33\% & 90.33\% & 83.33\% & 47.78\% & 76.45\% & 511.83  & 3206.63 & 6062.81 & 11224.51 & 5251.44 & -0.47 \\
\rowcolor{green!5}
ShorterBetter        & 79.33\% & 72.33\% & 66.67\% & 37.78\% & 64.03\% & 140.59  & 585.57  & 1613.66 & 4701.88  & 1760.43 & -1.45 \\
\rowcolor{green!5}
LC-R1             & 85.43\% & 88.67\% & 85.00\% & 42.22\% & 75.33\% & 449.44  & 1374.06 & 2788.19 & 6371.22  & 2745.73 & -0.22 \\
\rowcolor{green!5}
Eff$_{Rea}$               & 91.67\% & 91.33\% & 88.33\% & 53.33\% & 81.17\% & 1082.18 & 2700.98 & 4649.27 & 10706.94 & 4784.84 & 0.18 \\
\rowcolor{green!5}
LASER$_{DE}$             & 93.00\% & 91.33\% & 88.33\% & 51.11\% & 80.94\% & 974.57 & 1795.42 & 2766.08 & 5985.25 & 2880.33 & 0.44 \\
\rowcolor{green!5}
AutoTHINK         & 92.83\% & 93.67\% & 88.33\% & 47.78\% & 80.65\% & 1121.24 & 2449.16 & 3846.43 & 8303.59  & 3930.10 & 0.25 \\
\rowcolor{blue!5}
CoT-Valve          & 90.00\% & 80.33\% & 65.00\% & 20.00\% & 63.83\% & 328.73  & 1499.40 & 1940.33 & 4177.39  & 1986.46 & -1.51 \\
\rowcolor{blue!5}
C3oT               & 92.50\% & 92.00\% & 87.50\% & 51.11\% & 80.78\% & 1475.27 & 3805.73 & 6820.08 & 12884.91 & 6246.50 & -0.09 \\
\rowcolor{blue!5}
S$^{3}$-CoT         & 91.17\% & 92.00\% & 90.83\% & 51.11\% & 81.28\% & 1182.04 & 2833.27 & 5715.74 & 12217.53 & 5487.14 & 0.09 \\
\rowcolor{blue!5}
S$^{3}$-CoT$^{sc}$  & 91.67\% & 90.67\% & 87.50\% & 50.00\% & 79.96\% & 1149.86 & 3016.64 & 5654.80 & 12167.95 & 5497.31 & -0.07 \\
\bottomrule
\end{tabular}
\vspace{-0.2cm}
\caption{Evaluation on math benchmarks under Qwen2.5$_{7B}$ and DeepSeek-R1$_{7B}$.}
\vspace{-0.5cm}
\label{tab:main_exp_math_qw25_dsqw}
\end{table*}

\begin{table*}[t]
\centering
\small
\setlength{\tabcolsep}{2.7pt}
\begin{tabular}{lcccc|cccc|c}
\toprule
\multirow{2}{*}{Method} &
\multicolumn{4}{c|}{Accuracy$\uparrow$} &
\multicolumn{4}{c|}{Length$\downarrow$} &
\multirow{2}{*}{AES$\uparrow$} \\
\cmidrule(lr){2-5}\cmidrule(lr){6-9}
& MedQA & MedMCQA & BULLET & AVG. & MedQA & MedMCQA & BULLET & AVG. & \\
\midrule
\multicolumn{10}{c}{\textbf{Qwen2.5$_{7B}$}} \\
% \midrule
Standard$_p$        & 40.83\% & 56.00\% & 26.83\% & 41.22\% & 498.30 & 345.58 & 514.86 & 452.91 & -- \\
Efficient$_p$       & 49.17\% & 60.33\% & 31.67\% & 47.06\% & 118.43 & 62.23  & 107.11 & 95.92  & 1.50 \\
\rowcolor{blue!5}
TokenSKIP            & 45.67\% & 51.33\% & 23.17\% & 40.06\% & 461.66 & 335.16 & 464.99 & 420.60 & -0.21 \\
\rowcolor{blue!5}
CoT-Valve          & 55.00\% & 61.00\% & 40.33\% & 52.11\% & 564.33 & 411.82 & 584.32 & 520.16 & 1.17 \\
\rowcolor{blue!5}
C3oT              & 55.33\% & 61.50\% & 38.33\% & 51.72\% & 430.10 & 297.32 & 429.26 & 385.56 & 1.42 \\
\rowcolor{blue!5}
S$^{3}$-CoT         & 54.50\% & 60.17\% & 34.83\% & 49.83\% & 302.19 & 192.57 & 309.47 & 268.07 & 1.45 \\
\rowcolor{blue!5}
S$^{3}$-CoT$^{sc}$  & 52.67\% & 59.33\% & 33.83\% & 48.61\% & 304.07 & 195.01 & 306.22 & 268.43 & 1.30 \\
\midrule
\multicolumn{10}{c}{\textbf{DeepSeek-R1$_{7B}$}} \\
% \midrule
Standard$_p$        & 38.67\% & 37.33\% & 30.83\% & 35.61\% & 1865.66 & 1339.95 & 2169.36 & 1791.66 & -- \\
Efficient$_p$       & 36.83\% & 34.33\% & 27.50\% & 32.89\% & 1362.70 & 746.79  & 1297.02 & 1135.50 & -0.40 \\
\rowcolor{green!5}
ShorterBetter       & 37.17\% & 36.00\% & 30.47\% & 34.54\% & 692.33  & 357.91  & 716.78  & 589.00  & 0.37 \\
\rowcolor{green!5}
LC-R1                & 35.83\% & 37.50\% & 29.17\% & 34.17\% & 1310.58 & 672.80  & 1248.25 & 1077.21 & -0.01 \\
\rowcolor{green!5}
Eff$_{Rea}$                 & 37.00\% & 36.33\% & 31.83\% & 35.06\% & 1736.26 & 948.04  & 1651.76 & 1445.35 & 0.04 \\
\rowcolor{green!5}
LASER$_{DE}$                  & 38.33\% & 37.67\% & 29.33\% & 35.11\% & 1354.27 & 868.28  & 1286.06 & 1163.54 & 0.21 \\
\rowcolor{green!5}
AutoTHINK            & 39.50\% & 38.33\% & 31.33\% & 36.39\% & 1742.88 & 1117.90 & 1692.67 & 1517.82 & 0.26 \\
\rowcolor{blue!5}
CoT-Valve            & 28.00\% & 30.83\% & 24.67\% & 27.83\% & 1153.82 & 1676.99 & 918.82  & 1249.88 & -1.88 \\
\rowcolor{blue!5}
C3oT                 & 36.50\% & 36.83\% & 32.67\% & 35.33\% & 1832.41 & 1265.82 & 1802.02 & 1633.42 & 0.01 \\
\rowcolor{blue!5}
S$^{3}$-CoT         & 39.50\% & 36.67\% & 30.17\% & 35.45\% & 1648.16 & 1024.41 & 1608.56 & 1427.04 & 0.16 \\
\rowcolor{blue!5}
S$^{3}$-CoT$^{sc}$  & 40.00\% & 38.50\% & 28.17\% & 35.56\% & 1647.53 & 960.44  & 1555.83 & 1387.93 & 0.21 \\
\bottomrule
\end{tabular}
\vspace{-0.2cm}
\caption{Evaluation on medical benchmarks under Qwen2.5$_{7B}$ and DeepSeek-R1$_{7B}$.}
\label{tab:main_exp_med_qw25_dsqw}
\vspace{-0.2cm}
\end{table*}

\vspace{-0.1cm}
\subsection{SFT Method}

% 对于SFT，我们采用了类人的双认知系统和逐渐压缩策略。
% 双系统已经被证明是人类的一种认知模式，即可以给出快速的思维（系统1），也可以给出慢思考（系统2）。
% 如图1所示，我们的研究通过引入系统1提示和系统2提示来建立这一认知系统，其中在系统1提示下模型的学习目标是压缩后的CoT，而在系统2提示下其目标是原始回复。
% 与此同时，我们在分析中发现从采样数据中选取最短的CoT来指导SFT，会造成过度压缩损害LLM的表现。
% 因此，我们采取逐渐压缩的方式，即在训练过程中逐渐扩展len-R的压缩比分布范围，从[0.9,1.0]逐渐扩展到[0.0,1.0]。
% 每次扩展我们都按照给定的压缩比分布范围，从训练数据中进行采样，尽可能按照len-R满足均匀分布。
% 在压缩过程中，我们遵化标准的SFT过程，通过一个小的评估集测试性能，并采取早停的设置策略。

For SFT, our study adopts a dual-cognitive system and a progressive compression curriculum. 
Dual-cognitive theory suggests that human cognition comprises both fast thinking (System 1) and slow reasoning (System 2) \citep{evans2008dual}. 
As illustrated in Fig.~\ref{fig_overall_framework}, we instantiate this framework by introducing the System 1 prompt and the System 2 prompt. 
Under the System 1 prompt, the learning objective is a compressed CoT, whereas under the System 2 prompt, the learning objective is the initial response.
Meanwhile, our analysis (Sec.~\ref{sec_ana_shortest}) shows that using the shortest CoTs to supervise SFT typically leads to over-compression, which significantly degrades LLM performance.
To mitigate this issue, we adopt a progressive compression curriculum.
As shown in Fig.~\ref{fig_overall_framework}, during training, we progressively expand the distribution of Len-R from ([0.9, 1.0]) to ([0.0, 1.0]) with a step size of 0.1. 
At each iteration, we sample data to make Len-R as close to uniformly distributed as possible.
Throughout compression, we follow a standard SFT pipeline, evaluate on a small validation set to select checkpoints.

\vspace{-0.1cm}
\subsection{Experiment settings}

\paragraph{Training Data and LLMs.}
For training data, our study only uses the variable-length CoT data sampled from GSM8K as described in Sec.~\ref{sec_sample}.
We refer to an answer-verification based method as \textbf{S$^{3}$-CoT}, and the self-consistency verification based method as \textbf{S$^{3}$-CoT$^{sc}$}.
We conduct experiments on both general LLMs (Qwen2.5$_{7B}$/LLaMA3$_{8B}$) and R1-style LLMs (Deepseek-R1$_{7B}$/Qwen3-Think$_{4B}$). 
On Qwen2.5$_{7B}$ and Deepseek-R1$_{7B}$, we compare against state-of-the-art baselines, while on LLaMA3$_{8B}$ and Qwen3-Think$_{4B}$ we further demonstrate the adaptability of our method.

\paragraph{Baselines and Settings.}
For baselines, our study considers three families of methods: prompt control (Standard$_{p}$ and Efficient$_{p}$ \citep{renze2024benefits}), SFT-based (TokenSkip \citep{xia2025tokenskip}, C3oT \citep{kang2025c3ot}, and CoT-Valve \citep{ma2025cot}), and RL-based (ShorterBetter \citep{yi2025shorterbetter}, LC-R1 \citep{cheng2025optimizing}, Eff$_{Rea}$ \citep{arora2025training}, LASER$_{DE}$ \citep{liu2025laser}, and AutoTHINK \citep{tu2025learning}).
Their implementation details are in Appendix~\ref{app_baselines}.
Notably, RL-based methods are typically trained on DeepScaleR-Preview \citep{luo2025deepscaler}, a composite dataset of 40K instances covering AIME, AMC, MATH, etc. 
Their training process is highly compute-intensive, typically requiring 8×NVIDIA 80GB H100 GPUs.
For our method, we adopt the LoRA framework \citep{hu2021lora} for SFT, and the LoRA hyperparameters are set to $r=8$ and $\alpha=16$.
Our method just requires 2×NVIDIA 80GB A100 GPUs.
For decoding, we follow the official generation settings, and the maximum generation length is set to 65,536.

% During training, we set the learning rate to 2e-5 and trained for 5 epochs per distribution segment. 

% We used early stopping with a patience of 3 expansions and selected checkpoints based on performance on GSM8K validation.

% Notably, R1-style LLM outputs often exceed 16,384 tokens, and truncating generations at 16,384 is an unfavorable setting for base LLMs.
% Therefore, we increase the maximum generation length from the commonly used 16,384 in prior work to 65,536. 

% For prompt constraints, Standard$_{p}$ uses standard prompting to elicit reasoning, whereas Efficient$_{p}$ imposes a conciseness constraint (``Be concise.'' ).
% For SFT-based methods, we include TokenSkip, C3oT, and CoT-Value:
% TokenSkip releases weights trained on Qwen2.5$_{7B}$.
% For C3oT, we follow their settings, which prompt GPT-4o to remove redundant content.
% And for CoT-Value, we adopt their provided variable-length GSM8K data sampled based on QwQ-32B.
% We use the obtained data to guide SFT to re-implement their work. 
% For RL-based methods, we consider ShorterBetter, LC-R1, LASER, and AutoTHINK, which release weights on Deepseek-R1$_{7B}$. 

\paragraph{Evaluation Data and Metrics.}

For evaluation data, we follow mainstream evaluation on math benchmarks, including GSM8K, MATH \citep{lightman2023let}, AMC23 \citep{amc23_mathai}, and AIME24 \citep{aime24_mathai}. 
Considering that RL-based methods are trained on a math mixture that may overlap with test distributions, we further assess generalization on cross-domain medical benchmarks, including MedQA \citep{jin2021disease}, MedMCQA \citep{pal2022medmcqa}, and BULLET \citep{chen2025benchmarking}.
A detailed description of evaluation data can be found in Appendix.~\ref{app_eval_data}.
For the evaluation metric, we report accuracy averaged over three random responses, along with the corresponding average response token length.
Moreover, we adopt the AES metric proposed in prior work \citep{luo2025o1} to quantify the length-accuracy trade-off, computed as:
\begin{equation}
\small
\mathrm{AES}=
\begin{cases}
\omega \cdot \Delta \mathrm{Length} + \beta \cdot \lvert \Delta \mathrm{Acc} \rvert, & \text{if } \Delta \mathrm{Acc} \ge 0,\\
\omega \cdot \Delta \mathrm{Length} - \gamma \cdot \lvert \Delta \mathrm{Acc} \rvert, & \text{if } \Delta \mathrm{Acc} < 0.
\end{cases}
\end{equation}
where 
$\Delta \mathrm{Length}$ represents the difference from the response length under Standard$_{p}$, $\Delta \mathrm{Acc}$  represents the difference from the accuracy under Standard$_{p}$, 
and $\omega$, $\beta$, and $\gamma$ are set to 1, 5, and 10, respectively.

\begin{table*}[t]
\centering
\small
\setlength{\tabcolsep}{2.7pt}
\begin{tabular}{lccccc|ccccc|c}
\toprule
\multirow{2}{*}{Method} & \multicolumn{5}{c|}{Accuracy$\uparrow$} & \multicolumn{5}{c|}{Length$\downarrow$} & \multirow{2}{*}{AES$\uparrow$}  \\
\cmidrule(lr){2-6}\cmidrule(lr){7-11}
& GSM8K & MATH & AMC23 & AIME24 & AVG.
& GSM8K & MATH & AMC23 & AIME24 & AVG.
&  \\
\midrule
\multicolumn{12}{c}{\textbf{LLaMA3$_{8B}$}} \\
Standard$_{p}$        & 78.83\% & 47.67\% & 20.00\% & 3.33\% & 37.46\% & 245.06 & 605.28 & 857.21 & 1314.42 & 755.49 & -- \\
Efficient$_{p}$  & 68.67\% & 43.00\% & 24.17\% & 3.33\% & 34.79\% & 109.61 & 422.35 & 708.14 & 937.38  & 544.37 & -0.43\\
\rowcolor{blue!5}
S$^{3}$-CoT       & 80.17\% & 50.33\% & 25.00\% & 4.44\% & 39.99\% & 179.42 & 445.30 & 734.82 & 898.33  & 564.47 & 0.59 \\
\rowcolor{blue!5}
S$^{3}$-CoT$^{sc}$ & 79.67\% & 49.33\% & 22.50\% & 4.44\% & 38.99\% & 176.11 & 496.30 & 699.45 & 1036.81 & 602.17 & 0.41 \\
\midrule
\multicolumn{12}{c}{\textbf{Qwen3-Think$_{4B}$}} \\
Standard$_{p}$        & 96.00\% & 93.00\% & 99.17\%  & 82.22\% & 92.60\% & 1507.94 & 5573.16 & 10956.53 & 21009.29 & 9761.73 & -- \\
Efficient$_{p}$  & 94.33\% & 91.33\% & 96.67\%  & 76.67\% & 89.75\% & 812.05  & 4150.17 & 9344.19  & 18074.15 & 8095.14 & -0.76 \\
\rowcolor{blue!5}
S$^{3}$-CoT       & 94.83\% & 92.33\% & 100.00\% & 76.67\% & 90.96\% & 1029.56 & 4102.99 & 9180.60  & 17284.10 & 7899.31 & -0.41 \\
\rowcolor{blue!5}
S$^{3}$-CoT$^{sc}$ & 95.00\% & 92.00\% & 98.33\%  & 76.67\% & 90.50\% & 1061.49 & 4249.95 & 9162.59  & 17308.79 & 7945.71 & -0.54 \\
\bottomrule
\end{tabular}
\vspace{-0.1cm}
\caption{Evaluation on math benchmarks under LLaMA3$_{8B}$ and Qwen3-Think$_{4B}$.}
\vspace{-0.4cm}
\label{tab:main_exp_math_ll3_qw3}
\end{table*}

\vspace{-0.1cm}
\subsection{Main Results}

\paragraph{Compare with strong baselines.}

%表1报告了不同方法在数学基准上的评估性能，而表2报告了他们在医学基准上的性能。
% 基于提示约束的方法虽然在压缩长度方面具有一定作用，但通常会大幅损耗任务性能。
% 这凸显了通过SFT-based和RL-based方法内化高效推理能力的重要性。
% 对于数学基准上的评测上，在qwen2.5$_{7B}$上相比于standardp，我们的方法可以在性能上略有提升的同时将长度压缩平均150个tokens（大约为初始的百分之20左右）。
% 相比于各种强基线，我们的方法在性能上可以接近最优的性能，并且在AES指标上达到最优，这表明我们的方法更加均衡。
% 在DeepSeek-R1$_{7B}$上相比于standardp，我们的方法在性能上有一个轻微的下降同时将平均长度压缩1100个tokens（大约为初始的百分之17左右）。
% 相比于SFT-based方法（蓝色覆盖的区域），我们的方法在AES指标显著更优。
% 其中，CoT-valve由于过度压缩造成性能大幅下降，而C3oT尽管维持了任务性能但在长度压缩方面并不显著。
% 相比于RL-based方法（绿色覆盖的区域），在性能方面，我们的方法显著优于Shorterbetter和LC-R1，并与LASER和AutoTHINK持平。
% 但在长度压缩方面，相比于LASER和AutoTHINK，我们的方法似乎存在一定的劣势。
% 对于这一点，我们担心这背后的原因是由于RL-based方法在训练期间见过同分布的数学测试数据所造成的。
% 一方面，我们观测到在与我们训练同分布的测试集GSM8K上，我们方法在长度压缩方面是可以与LASER和AutoTHINK相持平的。
% 另一方面，为了更加公平的比较，我们在一个未知分布下即医学基准下进行了进一步的评估。

% 如表2所示，在qwen2.5$_{7B}$上我们注意到一个有趣的现象：各类压缩方法在医学基准测试上性能几乎均有一个显著的提升，这体现了压缩思维链的重要性。
% 这一现象在先前的研究中被解释过，即冗余的思维链可能会由于噪声的积累造成推理路径的跑偏。
% 对于$qwen2.5$_{7B}$，在性能上我们的方法接近最优，而在长度方面可以压缩约180个tokens（约为初始长度的40%）。
% 相比于各种强基线，我们的方法仍然更加均衡，即在AES指标上保持较高的水准。
% 对于$DeepSeek-R1$_{7B}$, 具有竞争力的基线仍然是LASER和AutoTHINK，而对于其他基线，我们的方法展现出显著的优势。
% 而与LASER和AutoTHINK相比，我们的方法在性能和压缩长度上几乎都是持平的，并不像先前在压缩长度方面存在显著的劣势。
% 这一结果可以验证我们的猜测，即在数学基准上的劣势是由于训练与测试集同分布造成的。

% 从整体上看，相比于sft-based方法，我们方法的优势十分显著。
% 这不仅体现在构造监督数据时我们的方法不需要任何外部指导，而在性能和长度压缩方面，我们的方法均表现更好。
% 相比于RL-based方法，我们的方法在更少的训练资源下，可以达到一个可比的性能。
% 我们也需要声明的是，虽然在我们的研究中，我们将RL方法进行独立的比较。
% 但在现实中，SFT方法与RL方法本身是可以无缝集成的，即可以先通过SFT方法压缩，在通过RL方法进一步压缩。
% 我们也鼓励未来的研究进一步探索其潜力。

Tab.~\ref{tab:main_exp_math_qw25_dsqw} and ~\ref{tab:main_exp_med_qw25_dsqw} summarize results on math and medical benchmarks. 
We observe that prompt control (Efficient$_{p}$) can shorten CoT length but often causes significant accuracy drops, motivating SFT/RL-based methods to internalize efficient CoT.

On math benchmarks, our method improves the overall accuracy-length trade-off. 
For Qwen2.5$_{7B}$, compared to Standard$_p$, it reduces length by \textasciitilde150 tokens (\textasciitilde20\% of initial length) while slightly increasing accuracy. 
Compared to strong baselines, our method achieves near-best accuracy while attaining the best AES, indicating a more balanced trade-off.
For DeepSeek-R1$_{7B}$, our method compresses by \textasciitilde1,100 tokens (\textasciitilde17\% of initial length) with a small accuracy loss.
Compared with SFT-based baselines (blue-shaded region), it yields markedly better AES: CoT-Valve over-compresses and hurts accuracy, whereas C3oT preserves accuracy but offers limited compression.
Against RL-based baselines (green-shaded region), the accuracy of our method outperforms ShorterBetter and LC-R1 and is competitive with Eff$_{Rea}$, LASER$_{DE}$, and AutoTHINK, though behind them in length compression.
We guess that this gap stems from potential train–test distribution overlap in RL-based methods.
To enable a fair comparison, we further evaluate on medical benchmarks to assess generalization under a shifted distribution.

On medical benchmarks, for Qwen2.5$_{7B}$, our method can compress by \textasciitilde180 tokens (\textasciitilde40\% of initial length) while achieving near-best accuracy, and remains among the most balanced methods with a strong AES.
For DeepSeek-R1$_{7B}$, our method can compress by \textasciitilde300 tokens (\textasciitilde17\% of initial length) while maintaining accuracy.
The most competitive baselines remain Eff$_{Rea}$, LASER$_{DE}$, and AutoTHINK.
Compared with them, our method is essentially tied in accuracy and achieves comparable length compression—unlike the previously observed disadvantage.
This result supports our guess that the earlier compression gap may stem from distributional overlap between the training and test data.
Moreover, compared with other strong baselines, our method still shows clear advantages.

In aggregate, our method substantially outperforms SFT-based baselines without requiring external guidance, and matches RL-based baselines with fewer training resources. 
\emph{While we compare against RL-based methods separately, our method has the potential to be integrated with RL, serving as a warm-start (pre-training) stage before RL.
We leave a thorough exploration of this integration to future work.}

% our study addresses the limitation of supervision scarcity under SFT-based methods.

\begin{table}[t]
\centering
\small
\setlength{\tabcolsep}{5pt}
\begin{tabular}{lcccc}
\toprule
Dataset & \#Total & \#Retained & \#Correct & Acc. \\
\midrule
PRM12K & \multirow{2}{*}{2,000} & 1,427 & 1,395 & 97.76\% \\
MedQA  &   & 409  & 409  & 100.00\% \\
\bottomrule
\end{tabular}
\vspace{-0.1cm}
\caption{For PRM12K and MedQA datasets, the number and accuracy of samples retained by self-consistency verification under DeepSeek-R1$_{7B}$.}
\label{tab:self-consistency_ana}
\vspace{-0.3cm}
\end{table}

\begin{table}[t]
\centering
\small
\setlength{\tabcolsep}{1.8pt}
\begin{tabular}{lcccccc}
\toprule
\multirow{2}{*}{Method} & \multicolumn{3}{c}{PRM12K} & \multicolumn{3}{c}{MedQA} \\
\cmidrule(lr){2-4}\cmidrule(lr){5-7}
 & Acc.$\uparrow$ & Len.$\downarrow$ & AES$\uparrow$ & Acc.$\uparrow$ & Len.$\downarrow$ & AES$\uparrow$ \\
\midrule
Standard$_{p}$         & 81.90\% & 6564.23 & --    & 35.61\% & 1791.66 & --   \\
S$^{3}$CoT       & 80.67\% & 5206.07 & 0.06  & 35.14\% & 1278.56 & 0.15 \\
S$^{3}$-CoT$^{sc}$ & 79.33\% & 5570.84 & -0.16 & 35.37\% & 1238.28 & 0.24 \\
\bottomrule
\end{tabular}
\vspace{-0.1cm}
\caption{DeepSeek-R1$_{7B}$ trained on PRM12K and MedQA, evaluated on math and medical benchmarks, respectively. We report the average accuracy and length.}
\label{tab:results_other_data}
\vspace{-0.3cm}
\end{table}

\paragraph{Adaptability across various LLMs.}

% 我们的研究进一步在Llama3和Qwen3上指导实验，以验证我们的方法在不同模型上的适应性。
% 表1展示了数学基准上的评测结果，对于Llama3，我们的方法可以压缩大约160个tokens（约为初始长度的21%）同时提升1-2个点的准确率。
% 对于Qwen3，我们的方法可以压缩大约1800个tokens（约为初始长度的18%），并且有一个较小的性能下能下降。

% 结合先前的结果整体来看，对于通用模型来说，我们的方法不仅可以压缩思维链长度，还可以带来性能上的提升。
% 而对于R1风格的模型，我们的方法需要以一个较小的性能损失来实现思维链的压缩，但这目前也是所有方法的统一壁垒，需要在未来进一步解决。
%此外，我们需要强调S$^{3}$-CoT$^{sc}$代表了一种完全的模型自我进化，并且展现了很大的潜力。

We further evaluate our method on LLaMA3$_{8B}$ and Qwen3-Think$_{4B}$ to assess cross-model adaptability. 
Tab.~\ref{tab:main_exp_math_ll3_qw3} reports results on math benchmarks. 
For LLaMA3$_{8B}$, our method compresses by \textasciitilde160 tokens (\textasciitilde21\% of initial length) while improving accuracy by 1–2\% points.
For Qwen3-Think$_{4B}$, it compresses by \textasciitilde1,800 tokens (\textasciitilde18\% of initial length) with a small accuracy drop.
Combined with earlier results, these findings suggest that \emph{for general LLMs, our method can not only compress CoT length but also improve overall accuracy. 
But for R1-style LLMs, compression still incurs a slight accuracy trade-off, an open challenge shared by existing methods.}

Overall, our experiments comprehensively validate the effectiveness, generalization, and adaptability of our method.
In particular, \emph{S$^{3}$-CoT$^{sc}$, serving as a fully self-evolving variant, exhibits substantial potential}.

% length by (\sim)160 tokens ((\sim)21%) while improving accuracy by 1–2 points. On Qwen3, it shortens generations by (\sim)1,800 tokens ((\sim)18%) with a small accuracy drop. Combined with earlier results, these findings suggest that for general-purpose models, our method can both compress chains of thought and improve performance. For R1-style models, compression still incurs a modest performance trade-off—an open challenge shared by existing approaches. Finally, we note that S$^{3}$-CoT$^{sc}$ represents a fully self-evolving variant and exhibits substantial potential.

\vspace{-0.1cm}
\section{Analysis and Ablation}
\vspace{-0.2cm}

\begin{table}[t]
\centering
\small
\setlength{\tabcolsep}{2.2pt}
\begin{tabular}{lcccccc}
\toprule
\multirow{2}{*}{Method} & \multicolumn{3}{c}{DeepSeek-R1$_{7B}$} & \multicolumn{3}{c}{Qwen2.5$_{7B}$} \\
\cmidrule(lr){2-4}\cmidrule(lr){5-7}
 & Acc.$\uparrow$ & Len.$\downarrow$ & AES$\uparrow$ & Acc.$\uparrow$ & Len.$\downarrow$ & AES$\uparrow$ \\
\midrule
% Standard$_{p}$     & 81.90\% & 6564.23 & --    & 54.28\% & 673.14 & -- \\
S$^{3}$CoT    & 81.28\% & 5487.14 & 0.09  & 55.43\% & 522.29 & 0.33 \\
Short$_{only}$ & 74.89\% & 4495.83 & -0.54 & 50.97\% & 437.34 & -0.57 \\
\bottomrule
\end{tabular}
\vspace{-0.1cm}
\caption{Comparison against training with the shortest CoTs. We report the average accuracy and length.}
\vspace{-0.4cm}
\label{tab:results_w_short}
\end{table}

To provide deeper insight into our method, we present case studies in Appendix~\ref{app_case}. Here, we focus on answering the following two questions:

% To provide deeper insight into our method, we present case studies and show performance change on the validation set in Appendix~\ref{app_case} and ~\ref{app_vad_acc}. Here, we focus on answering the following two questions:

\subsection{Can our method generalize across various training datasets?}

To answer this question, we run additional experiments on DeepSeek-R1$_{7B}$ with PRM12K (math) \citep{lightman2023let} and MedQA (medical) \citep{jin2021disease} as training data. 
Based on the obtained variable-length direction and intervention settings as described in Sec.~\ref{sec_sample}, we sample 2,000 variable-length CoT instances per dataset. 
Fig.~\ref{app_len_r_dist} (in Appendix) shows the resulting Len–R distributions, confirming that our method can still produce variable-length traces across datasets. 
Tab.~\ref{tab:self-consistency_ana} reports the number and accuracy of samples retained by self-consistency verification.
Consistent with our earlier findings, this mechanism can help ensure the correctness of the sampled data: the retained MedQA samples even achieve 100\% accuracy. 
Tab.~\ref{tab:results_other_data} presents performance with our method trained on different datasets, and we observe that our method consistently yields substantial CoT length compression with minimal accuracy loss.

% However, the sampling efficiency is influenced by the underlying LLM capability.
% \citep{lightman2023let}\citep{jin2021disease}

% Table 5 reports self-consistency filtering results, which maintain high correctness; notably, the retained MedQA samples reach 100% accuracy. Table 6 shows that models trained on these samples still achieve substantial length compression with negligible performance loss.

% 为了回答这一问题，我们在数学PRM12K和医学MedQA数据集在deepseek模型上进一步指导实验。
% 这里，我们直接利用在第三节中识别的激活方向和探索的干预设置来采样变长数据，对于每个数据集我们采样2000条。
% 附录中的图1和图2展现了采样数据len-r的分布情况，即使在不同的数据集上我们的方法也可以采样出变长CoT数据。
% 表5提供了利用自一致性验证保留数据的数量以及准确率。
% 对于不同的数据集，自一致性机制可以保正采样数据的准确率，其中在medqa上保留的数据性能甚至高达100%。
% 表6展示了在不同数据集下利用我们方法训练模型的性能。
% 整体上看，我们的方法仍然可以显著压缩思维链长度，在几乎不损失任务性能的前提下。

\subsection{Why not use the shortest CoT samples as supervision?}\label{sec_ana_shortest}
% \vspace{-0.2cm}

% Some prior work often advocates supervising with the shortest possible CoT. 
% In contrast, we find that-even with self-sampled data—training on the shortest CoT leads to over-compression. 
% As shown in Table 7, selecting the shortest response per instruction as supervision can compress more tokens but substantially degrades accuracy on DeepSeek-R1$_{7B}$ and Qwen2.5$_{7B}$. 
% Since the accuracy-preserving compression limit is unknown a priori, the progressive compression curriculum is necessary.

Some prior work \citep{munkhbat2025self,kang2025c3ot} advocates supervising LLMs with the shortest possible CoT. 
In contrast, we find that—even with self-sampled data—training exclusively on the shortest CoT still leads to over-compression. 
As shown in Tab.~\ref{tab:results_w_short}, supervision with the shortest CoT achieves greater token compression but substantially degrades accuracy. 
This observation suggests that, since the accuracy-preserving compression limit is unknown a priori, a progressive compression curriculum is necessary.

 % Qwen2.5$_{7B}$

% 一些先前的研究鼓励使用尽可能短的监督数据来压缩模型。
% 然而，我们的研究发现即使数据来源于模型自身，使用最短的CoT进行监督也会造成过度压缩。
% 如表7所示，我们采样每条指令最短的回复来作为监督数据，并在A和B模型上指导实验。
% 我们可以观测到这样的操作尽管能压缩更多的tokens，但会显著破坏模型的任务性能。
% 结合我们附录中提供的验证集上的性能变化，由于我们无法先验的知道在不损害性能下模型的压缩极限，因此指导逐渐式压缩是十分必要的。

% \begin{table*}[t]
% \centering
% \small
% \setlength{\tabcolsep}{5pt}
% \begin{tabular}{lccccccccc}
% \toprule
% \multirow{2}{*}{Method} & \multicolumn{4}{c}{Accuracy} & \multicolumn{4}{c}{Length} & \multirow{2}{*}{AES} \\
% \cmidrule(lr){2-5}\cmidrule(lr){6-9}
% & QA & MCQA & BULLET & AVG. & QA & MCQA & BULLET & AVG. &  \\
% \midrule
% \multicolumn{10}{c}{\textbf{qwen3}} \\
% Standard$_{p}$        & 65.33\% & 48.67\% & 48.50\% & 54.17\% & 4233.23 & 3237.00 & 4749.95 & 4073.40 & - \\
% Efficient$_{p}$ & 63.67\% & 33.00\% & 49.67\% & 48.78\% & 2806.79 & 2005.34 & 2958.97 & 2590.37 & -1.92 \\
% \rowcolor{blue!5}
% S$^{3}$-CoT       & 75.00\% & 67.67\% & 55.67\% & 66.11\% & 3072.18 & 2386.84 & 3435.68 & 2964.90 & 0.93 \\
% \rowcolor{blue!5}
% S$^{3}$-CoT$^{sc}$ & 71.67\% & 68.00\% & 57.67\% & 65.78\% & 3141.31 & 2385.19 & 3421.59 & 2982.70 & 0.91 \\
% \bottomrule
% \end{tabular}
% \caption{ACC and Length with AES on QA/MEDQA/MEDBULLET.}
% \label{tab:qa_medqa_medbullet_aes_2dp}
% \end{table*}

\vspace{-0.1cm}
\section{Conclusion}
\vspace{-0.1cm}

In summary, our study proposes a self-sampling framework (S$^{3}$-CoT) for efficient CoT learning. 
We establish an end-to-end pipeline that guides how to sample high-quality, variable-length CoT from LLMs themselves, and extensive experiments demonstrate that our sampled data can enable efficient CoT LLMs. 
This line of exploration suggests an LLM-level capacity for self-evolution, and to the best of our knowledge, we are among the earliest teams to investigate this pathway.
In future work, we will better leverage sampled data to push beyond the length–performance Pareto frontier.

% In the future, we will explore how to more fully leverage sampled data to push beyond the Pareto frontier between length and performance.

% help compress CoT length with minimal loss in task performance.

% In our study, we leverage activation steering as a data-sampling tool to facilitate efficient CoT learning. 
% This line of exploration suggests an LLM-level capacity for self-evolution, and to the best of our knowledge, we are among the earliest teams to systematically investigate this pathway.

\section*{Limitations}
Our study alleviates the supervision data bottleneck in efficient CoT learning. 
However, how to more fully exploit the acquired variable-length data to push beyond the Pareto frontier between accuracy and length remains an open question. 
In particular, for R1-style LLMs, both our method and existing methods still face the challenge of slight accuracy degradation.
Moreover, SFT-based methods can naturally serve as a warm-start (pre-training) stage before RL.
Whether such an integration can yield additional performance gains is an important direction for future research.

% \begin{itemize}[leftmargin=*,noitemsep,topsep=0pt]

% \end{itemize}

% \section{Ethical Considerations}
% The offensive examples included in this paper serve solely as illustrations and are not intended to be instructive or to promote such content.

% \clearpage

\bibliography{custom_our}

% \clearpage
\appendix

\section{Detailed Descriptions of Existing Methods}\label{app_related}
Existing methods largely fall into three paradigms.
\textbf{Prompt-control} constrains reasoning at inference by injecting explicit length signals or structured formats: 
TALE curbs overlong chains by estimating the token budget \citep{han2025token},
Concise Thoughts leverages cues (such as ``Be concise.'') to bias LLMs toward shorter outputs \citep{nayab2024concise}, 
and Chain-of-Draft encourages minimal intermediate draft notes to retain problem structure with substantially reduced verbosity \citep{xu2025chain}.
\textbf{SFT-based methods} fine-tune LLMs with succinct CoT as supervision: C3oT obtains compressed traces with the help of GPT-4o and trains LLMs on them \citep{kang2025c3ot}.
CoT-Valve learns a length-controllable LoRA module on QwQ$_{32B}$ and scales the module strength to yield variable-length CoTs, which are used to distill other LLMs \citep{ma2025cot}.
\textbf{RL-based methods} optimize the length-accuracy trade-off with explicit reward signals \citep{luo2025o1}: 
ThinkPrune progressively tightens a hard token budget, penalizing trajectories that exceed the limit \citep{hou2025thinkprune}.
LASER applies length-based reward and difficulty-aware variants to discourage overthinking on easy instances \citep{liu2025laser}. 
ShorterBetter uses the shortest sample among multiple generations as a self-supervised target \citep{yi2025shorterbetter}. 
LC-R1 combines a global length reward with an additional compression reward to remove redundant thinking \citep{cheng2025optimizing}.
% According to our investigation, RL-based methods are typically employed only to R1-style LLMs.

\section{Implementation of Strong Baselines}\label{app_baselines}

For prompt control, Standard$_{p}$ uses standard prompting to elicit reasoning (``Please reason step by step, and put your final answer within \verb|\|boxed\{\}.'').
And Efficient$_{p}$ further imposes a conciseness constraint such as ``Be concise.''.
For SFT-based methods, we include TokenSkip, C3oT, and CoT-Valve:
TokenSkip has released weights\footnote{\url{huggingface.co/hemingkx/TokenSkip-Qwen2.5-7B-Instruct-GSM8K}} trained on Qwen2.5$_{7B}$.
For C3oT, we follow their settings, which prompt GPT-4o to remove redundant content.
And for CoT-Valve, we adopt their provided variable-length GSM8K\footnote{\url{huggingface.co/datasets/horseee/MixChain-Z-GSM8K}} data sampled from QwQ-32B.
Based on the obtained data, we guide SFT to re-implement their work according to their settings. 
Notably, since CoT-Valve is based on QwQ-32B, the sampled data are typically longer than the default responses of Qwen2.5$_{7B}$. Consequently, after training, Qwen2.5$_{7B}$ 's CoT tends to become longer rather than shorter. 
This outcome highlights the limitation of CoT-Valve: it does not universally compress CoT across all backbone LLMs.
For RL-based methods, we consider ShorterBetter\footnote{\url{huggingface.co/JingyangYi/SB\_DS7B\_alpha\_2/tree/main}}, LC-R1\footnote{\url{huggingface.co/zx10086/LCR1\_7B}}, Eff$_{Rea}$\footnote{\url{huggingface.co/daman1209arora/alpha\_0.1\_DeepSeek-R1-Distill-Qwen-7B}}, LASER$_{DE}$\footnote{\url{huggingface.co/hkust-nlp/Laser-DE-L4096-7B/tree/main}}, and AutoTHINK\footnote{\url{huggingface.co/SONGJUNTU/Distill-R1-7B-AutoThink-Stage3}}, which all release weights on DeepSeek-R1$_{7B}$. 
We reproduce their results by following the provided reasoning templates and decoding configurations.

\section{Description of Evaluation Data}\label{app_eval_data}

The description of our used evaluation data can be summarized as:
\begin{itemize}[leftmargin=*,noitemsep,topsep=0pt]
\item GSM8K: A grade-school math word problem benchmark designed to evaluate multi-step numerical reasoning and arithmetic skills.
\item MATH: A challenging competition-level mathematics dataset covering algebra, geometry, number theory, and calculus with step-by-step solution requirements.
\item AMC23: A benchmark derived from the AMC 2023 competition, consisting of multiple-choice problems that test advanced pre-college mathematical reasoning.
\item AIME24: A dataset based on the AIME 2024 exam, featuring short-answer problems that require precise symbolic reasoning and complex problem solving.
\item MedQA: A large-scale medical question answering dataset composed of USMLE-style multiple-choice questions assessing professional-level clinical knowledge.
\item MedMCQA: A medical multiple-choice QA benchmark sourced from Indian medical entrance exams, covering a broad range of clinical and basic medical topics.
\item BULLET: A recent medical reasoning benchmark focused on evaluating LLMs’ robustness and generalization in complex, evidence-intensive clinical decision scenarios.
\end{itemize} 
In our study, to control evaluation cost, we randomly subsample the test sets to form our final evaluation set. Specifically, we sample 200 and 100 instances from GSM8K and MATH, respectively, and use the full test sets for AMC23 and AIME24. 
For the medical benchmarks (MedQA, MedMCQA, and BULLET), we randomly sample 200 instances from each dataset.

\section{Case Study}\label{app_case}

\begin{figure*}[ht]
\centering
\subfigure[Responses of Qwen2.5$_{7B}$.]
{\includegraphics[scale=0.47]{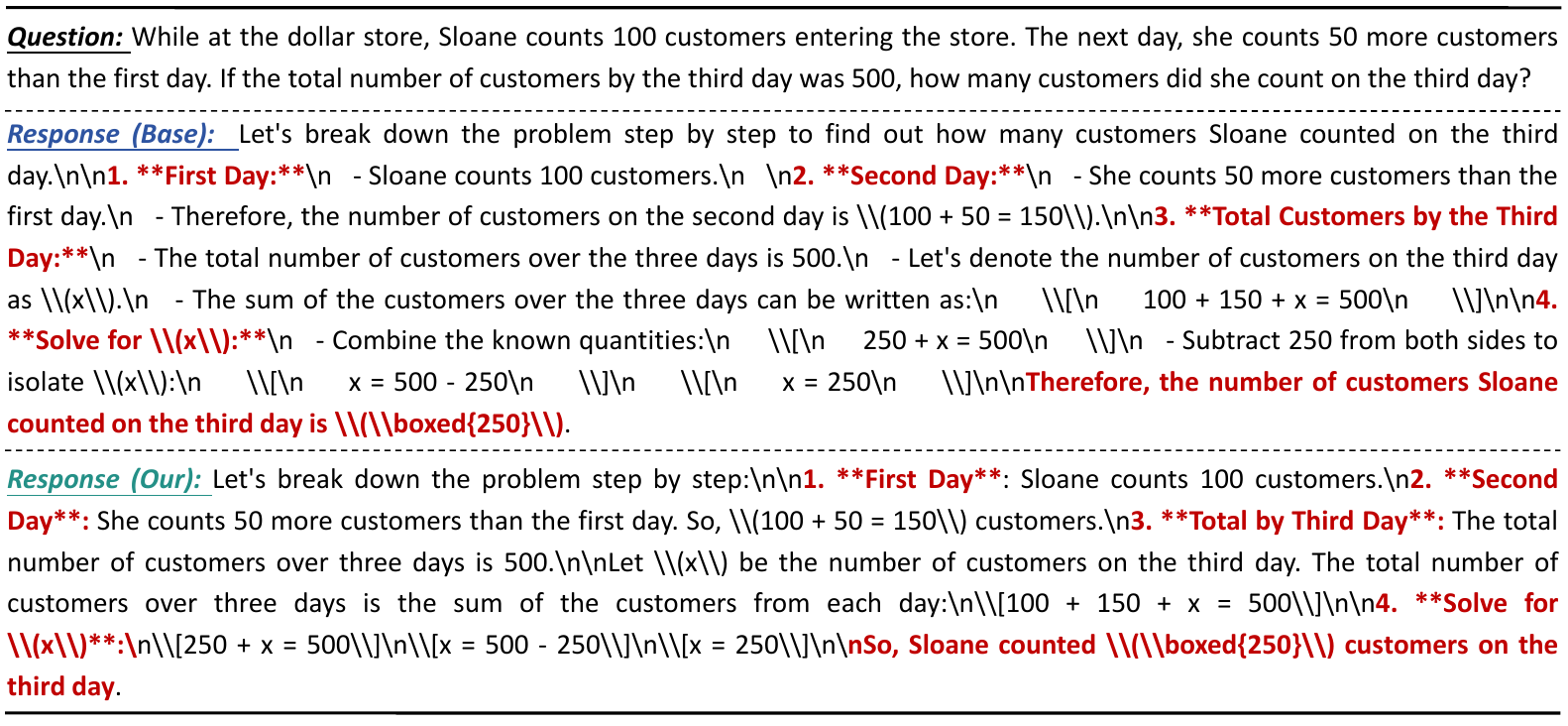}
}
% \hfill
\subfigure[Responses of DeepSeek-R1$_{7B}$.]
{\includegraphics[scale=0.47]{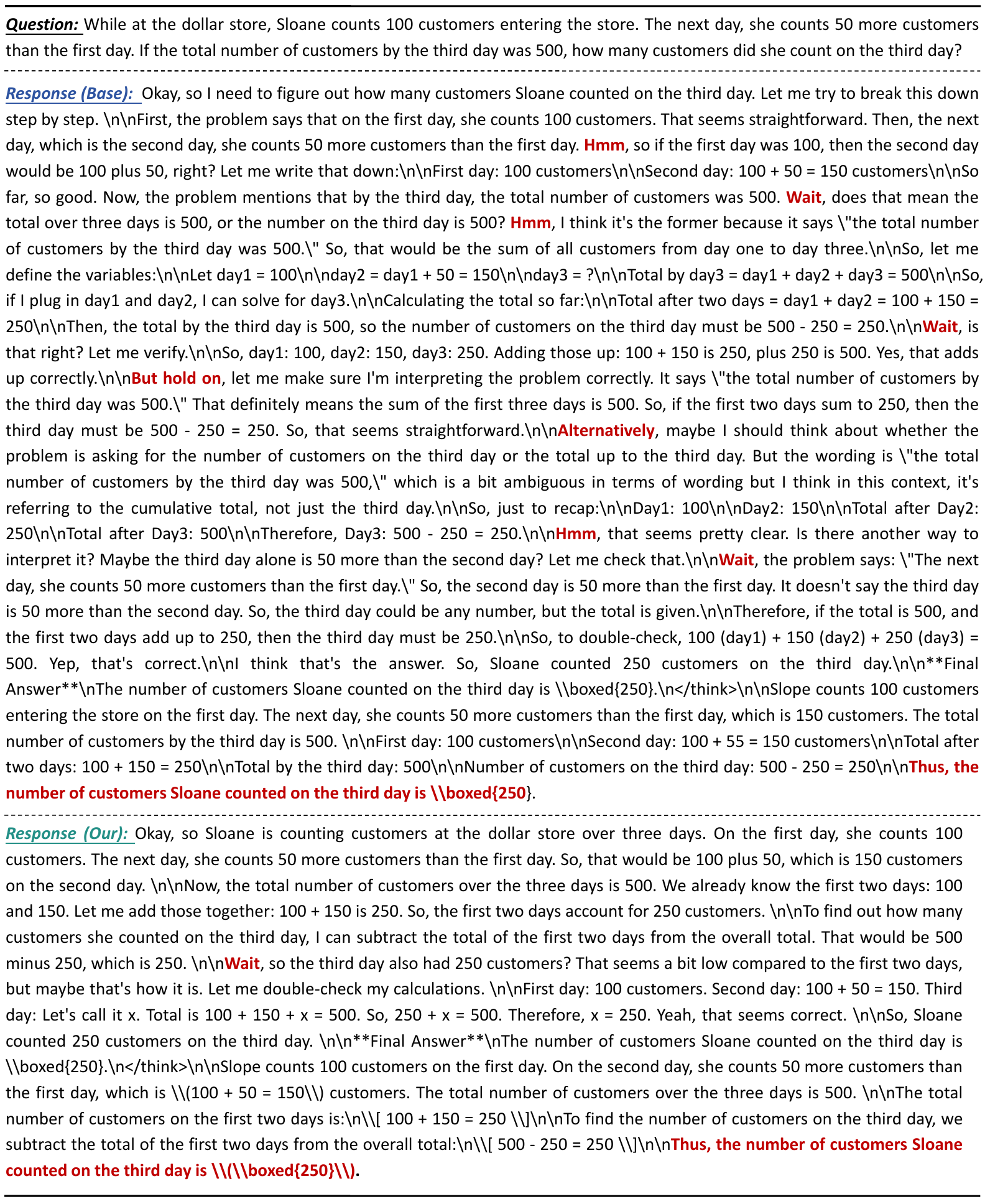}
}
\caption{Case study on Qwen2.5$_{7B}$ and DeepSeek-R1$_{7B}$. We highlight key reasoning steps and reflection steps in red.
 }
\label{app_fig_case}
\end{figure*}

As shown in Fig.~\ref{app_fig_case}, we present case studies on Qwen2.5$_{7B}$ and DeepSeek-R1$_{7B}$.
For Qwen2.5$_{7B}$, our method can remove redundant phrasing while preserving the core reasoning steps, yielding a more concise CoT without affecting correctness.
For DeepSeek-R1$_{7B}$, our method can further compress overly reflective behaviors.
For example, the base LLM performs eight rounds of reflection when answering the given question.
However, these reflections largely repeat the same viewpoint and amount to repeated self-verification.
In contrast, our method can retain LLMs' reflective ability while making the reflection more efficient and purposeful.

% \section{Tracing Performance on Validation Set}\label{app_vad_acc}

% In our study, we use 100 GSM8K samples as a validation set to guide checkpoint selection. 
% Fig.~\ref{app_dev_corr} and Fig.~\ref{app_dev_cons} show how the validation performance of various LLMs evolves under our S$^{3}$-CoT and S$^{3}$-CoT$^{sc}$, respectively.
% As described in Sec.~\ref{sec_main_exp}, we adopt a progressive compression curriculum, gradually expanding the Len-R distribution from ([0.9, 1.0]) to ([0.0, 1.0]) with a step size of 0.1. 
% We report the validation performance after each iteration, for a total of 10 iterations.
% We observe that the response length decreases steadily as compression increases.
% But the iteration at which peak accuracy occurs does not exhibit a consistent pattern across LLMs, suggesting that each LLM has a distinct accuracy-preserving compression limit. 
% Taking LLaMA3$_{8B}$ as an example, over-compression leads to a substantial accuracy drop, while other LLMs also show a mild degradation beyond their respective limits.
% Looking ahead, an important direction is to more effectively leverage the sampled variable-length data to push beyond the Pareto frontier between accuracy and length.

\begin{figure*}[ht]
\centering
\subfigure[Analysis on LLaMA3$_{8B}$.]
{\includegraphics[scale=0.65]{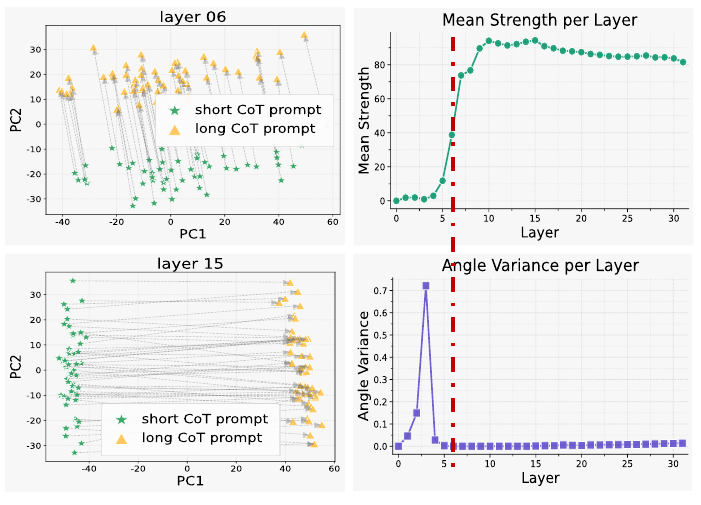}
}
% \hfill
\subfigure[Analysis on Qwen3-Think$_{4B}$.]
{\includegraphics[scale=0.65]{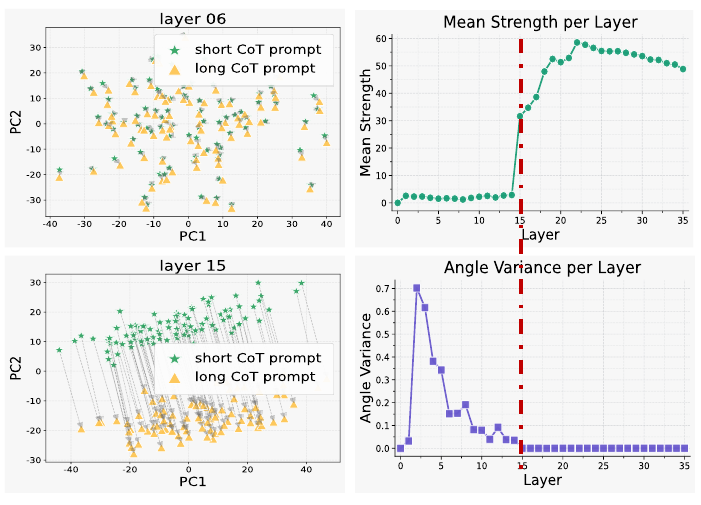}
}
\caption{Analysis of VL-D properties under LLaMA3$_{8B}$ and Qwen3-Think$_{4B}$. We provide PCA-based visualizations and quantify how the mean separation strength and angle variance metric vary across layers. Visualizations across all layers under various LLMs are in Fig.~\ref{app_vis_qw2.5},~\ref{app_vis_dsqw},~\ref{app_vis_ll3}, and ~\ref{app_vis_qw3}, respectively.}
\label{app_fig_ana}
\end{figure*}

\begin{figure*}[ht]
\centering
\subfigure[Analysis on LLaMA3$_{8B}$.]
{\includegraphics[scale=0.5]{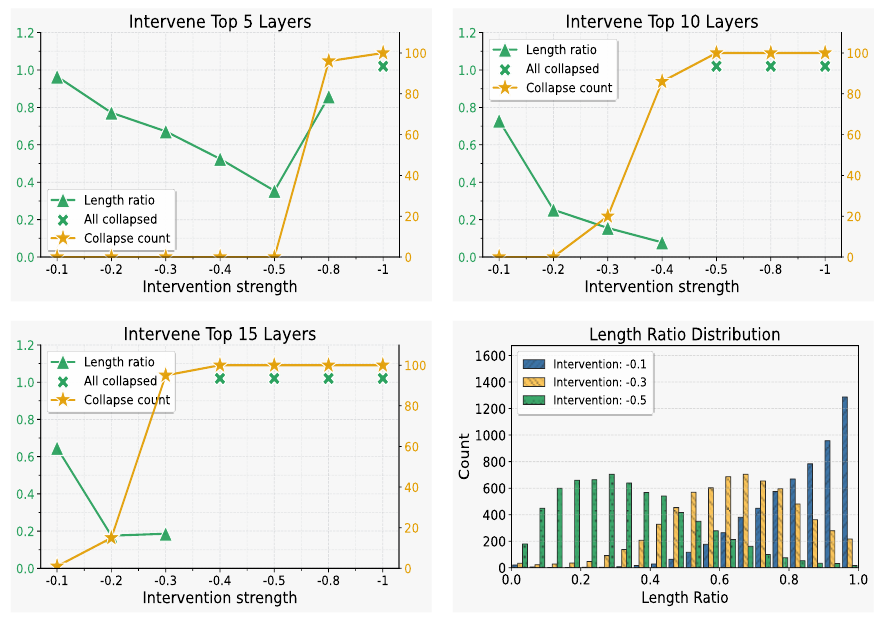}
}
% \hfill
\subfigure[Analysis on Qwen3-Think$_{4B}$.]
{\includegraphics[scale=0.5]{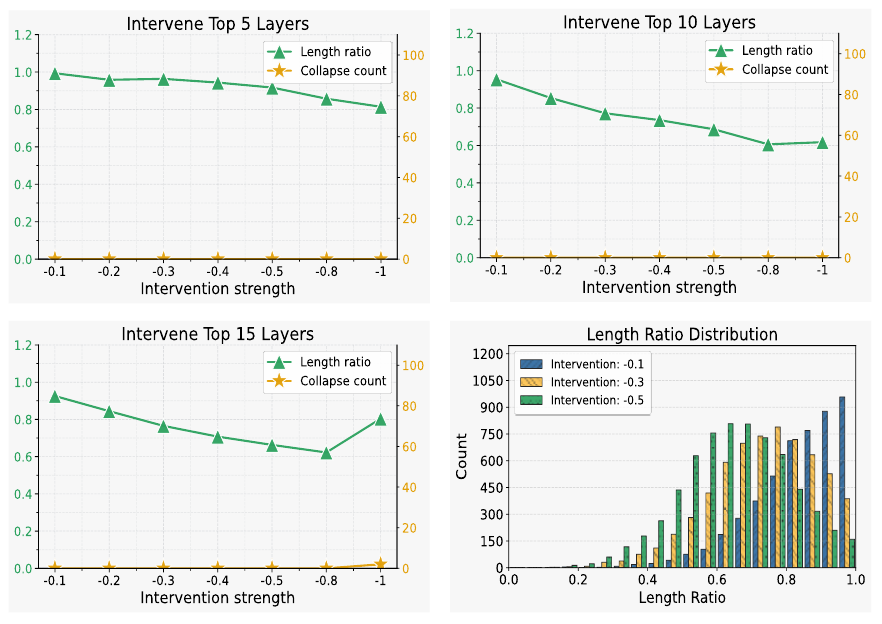}
}
\caption{Probe experiments on intervention layers and strength under LLaMA3$_{8B}$ and Qwen3-Think$_{4B}$. Green: average Len-R; Yellow: number of collapsed samples; Green ``×'': all samples collapse. Bottom-right: Len-R distribution under large-scale sampling. Results for other intervention settings are in Fig.~\ref{app_interve_other_sets}.}
\label{app_fig_ratio}
\vspace{-0.4cm}
\end{figure*}

\begin{figure*}[ht]
\centering
\subfigure[Analysis on Qwen2.5$_{7B}$.]
{\includegraphics[scale=0.5]{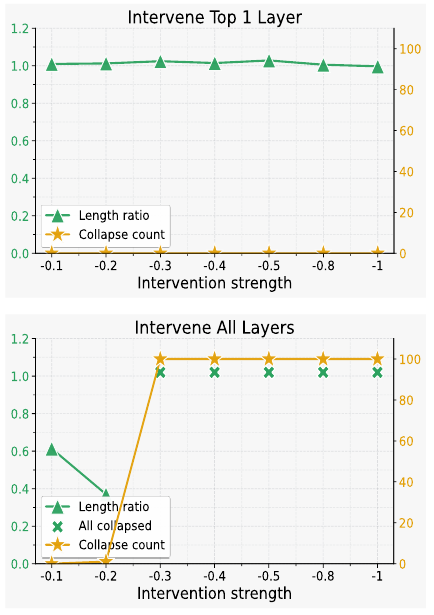}
}
\subfigure[Analysis on DeepSeek$_{7B}$.]
{\includegraphics[scale=0.5]{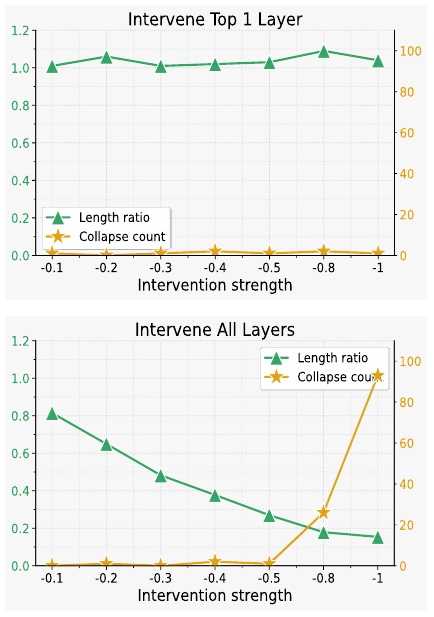}
}
\subfigure[Analysis on LLaMA3$_{8B}$.]
{\includegraphics[scale=0.5]{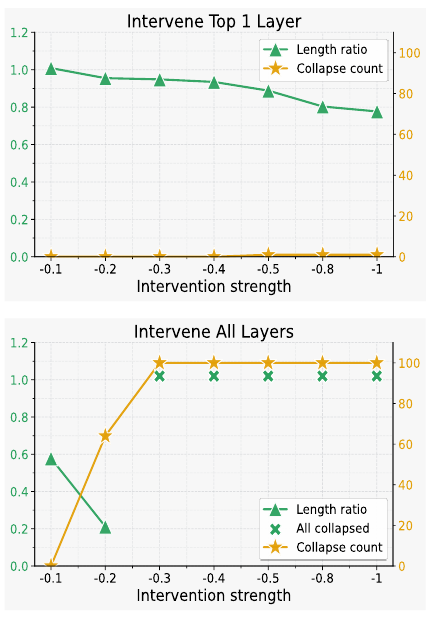}
}
\subfigure[Analysis on Qwen3$_{4B}$.]
{\includegraphics[scale=0.5]{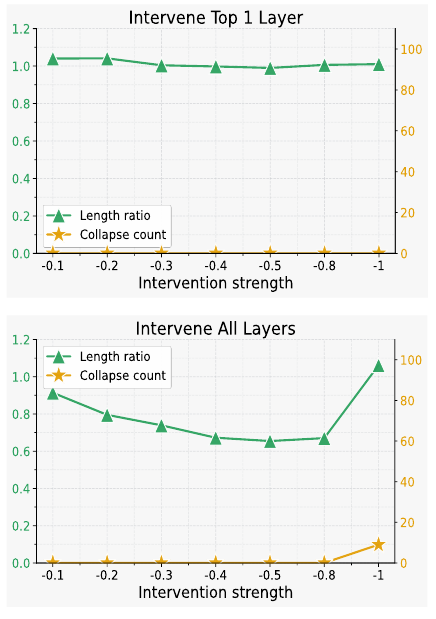}
}
\caption{Results for different intervention settings under various LLMs. Green: average Len-R; Yellow: number of collapsed samples; Green ``×'': all samples collapse.}
\label{app_interve_other_sets}
\vspace{-0.4cm}
\end{figure*}

\begin{figure*}[ht]
\centering
{\includegraphics[scale=0.57]{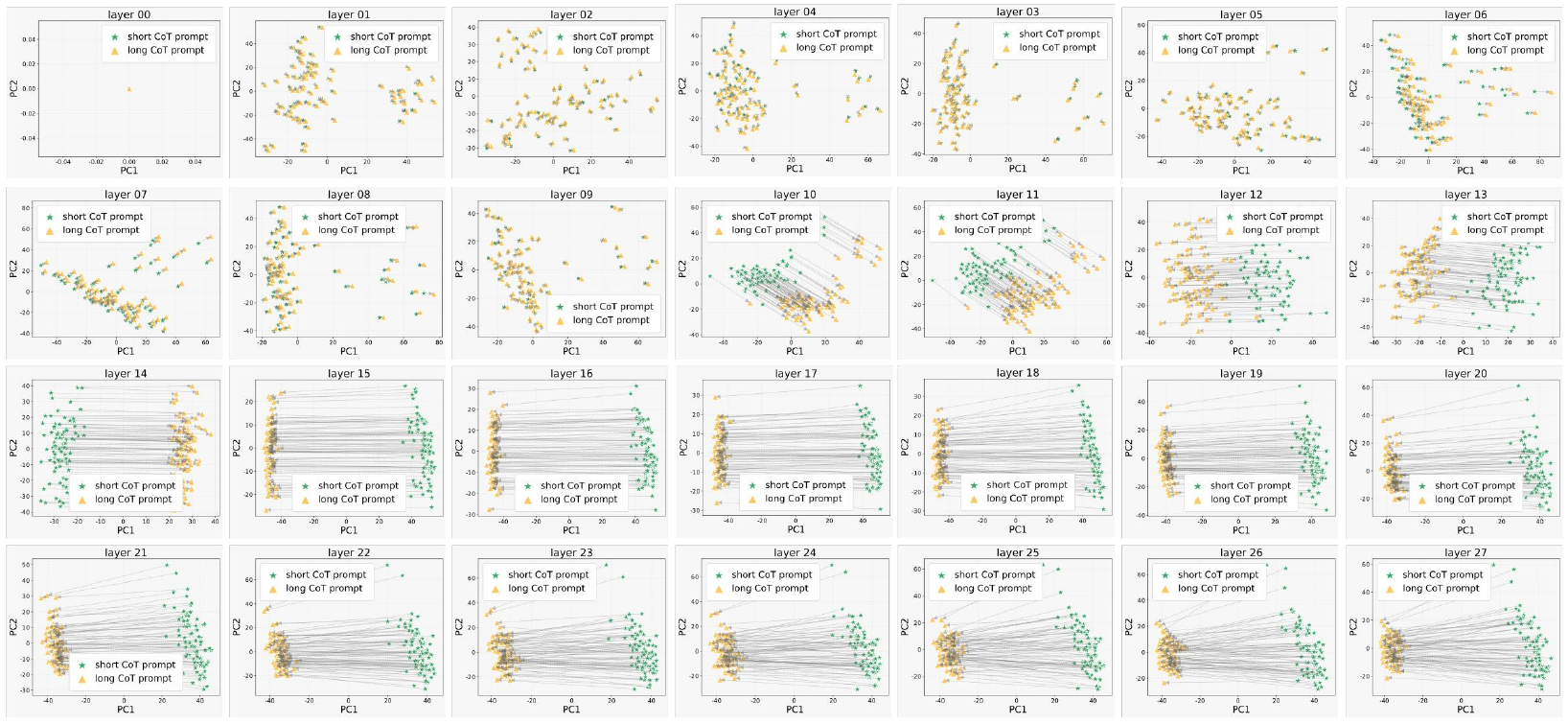}
}
\caption{Visualizations across all layers under Qwen2.5$_{7B}$.}
\label{app_vis_qw2.5}
\end{figure*}

\begin{figure*}[ht]
\centering
{\includegraphics[scale=0.58]{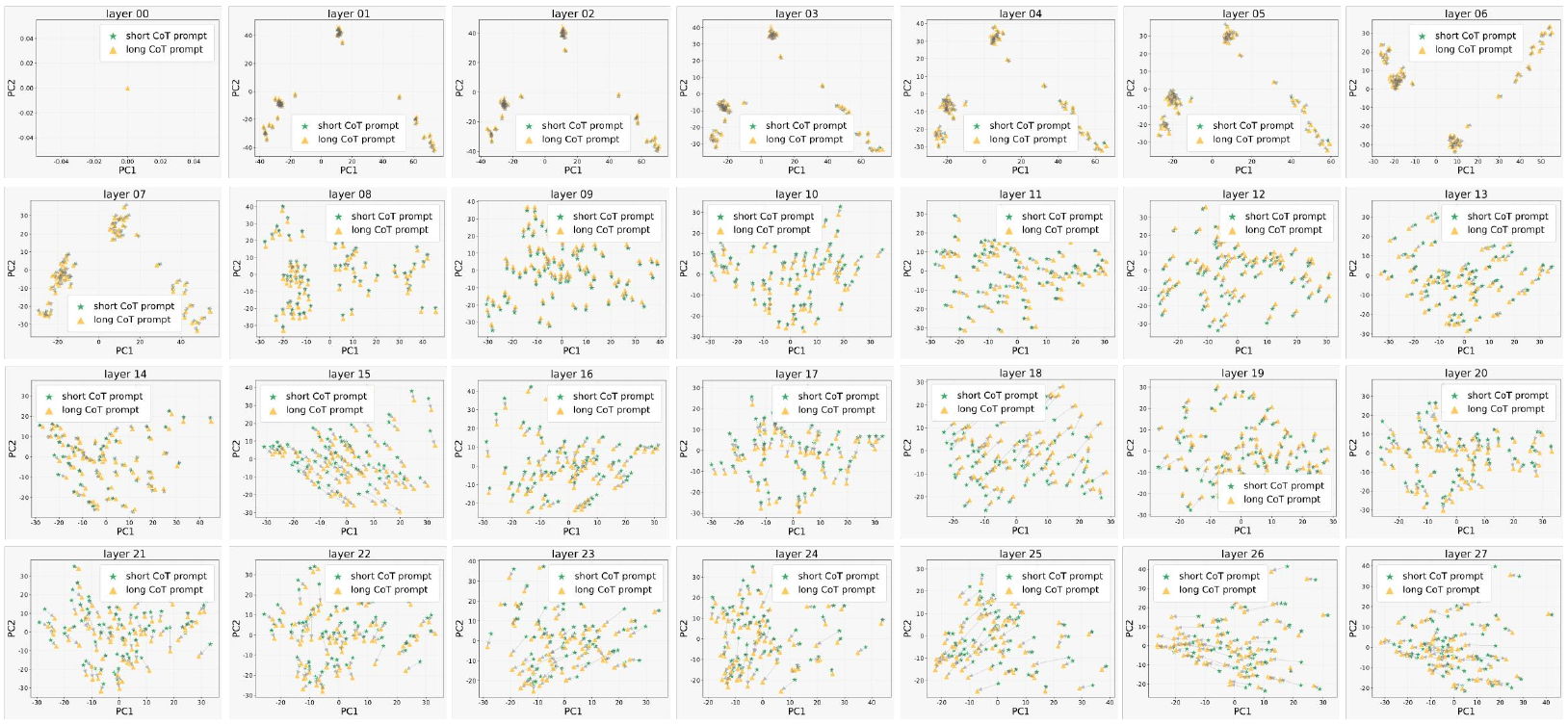}
}
\caption{Visualizations across all layers under DeepSeek-R1$_{7B}$.}
\label{app_vis_dsqw}
\end{figure*}

\begin{figure*}[ht]
\centering
{\includegraphics[scale=0.58]{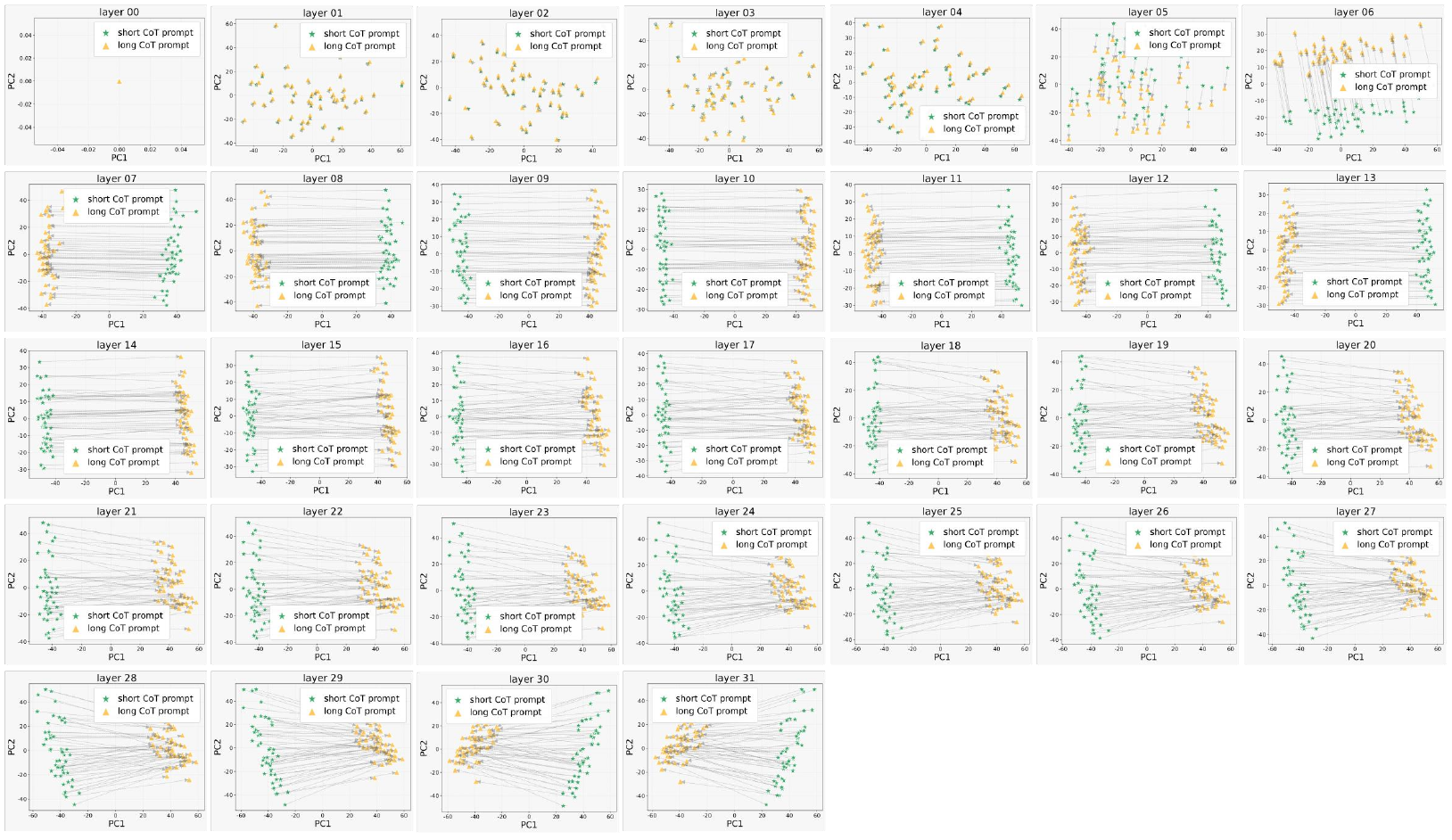}
}
\caption{Visualizations across all layers under LLaMA3$_{8B}$.}
\label{app_vis_ll3}
\end{figure*}

\begin{figure*}[ht]
\centering
{\includegraphics[scale=0.58]{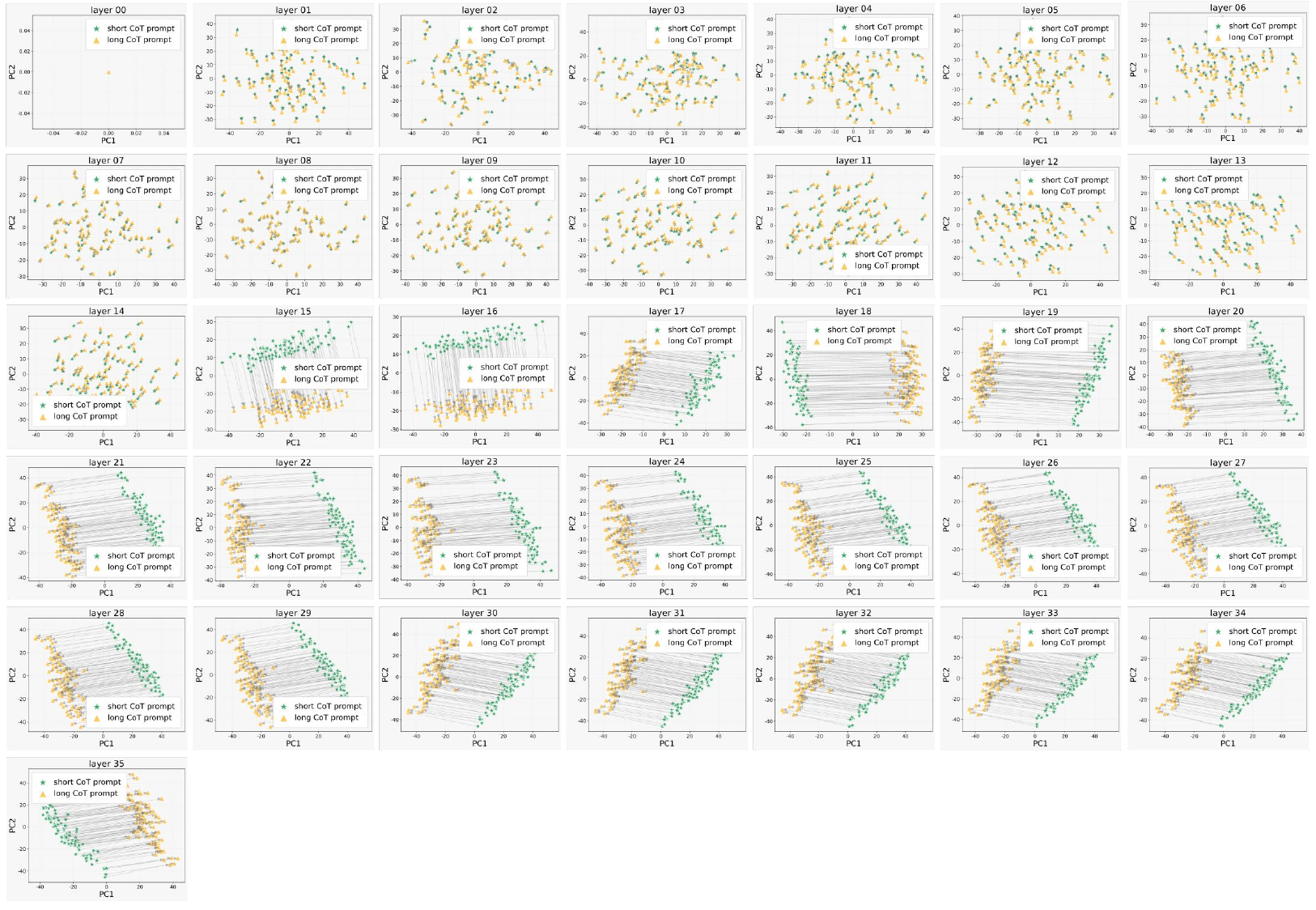}
}
\caption{Visualizations across all layers under Qwen3-Think$_{4B}$.}
\label{app_vis_qw3}
\end{figure*}

\begin{figure*}[ht]
\centering
\subfigure[Sampled from PRM12K.]
{\includegraphics[scale=0.7]{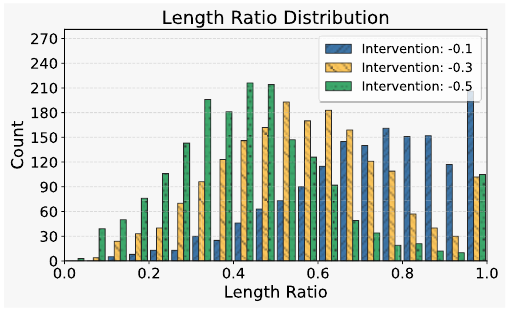}
}
% \hfill
\subfigure[Sampled from MedQA.]
{\includegraphics[scale=0.7]{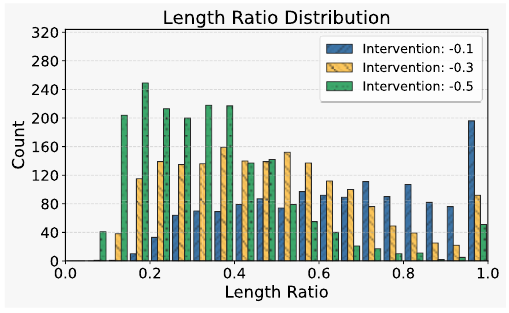}
}
\caption{Len-R distributions of sampled data from PRM12K and MedQA, respectively.
As the intervention strength increases, the overall distribution shifts left, indicating shorter CoT on average.}
\label{app_len_r_dist}
\end{figure*}

% \begin{figure*}[ht]
% \centering
% \subfigure[Analysis on Qwen2.5$_{7B}$.]
% {\includegraphics[scale=0.7]{Figure/dev_corr_qw2.5.pdf}
% }
% \subfigure[Analysis on DeepSeek-R1$_{7B}$.]
% {\includegraphics[scale=0.7]{Figure/dev_corr_dsqw.pdf}
% }
% \subfigure[Analysis on LLaMA3$_{8B}$.]
% {\includegraphics[scale=0.7]{Figure/dev_corr_ll3.pdf}
% }
% \subfigure[Analysis on Qwen3$_{4B}$.]
% {\includegraphics[scale=0.7]{Figure/dev_corr_qw3.pdf}
% }
% \caption{For each iteration, performance on the validation set under our S$^{3}$-CoT method. Green: Accuracy; Yellow: Length; Red Circle: Our selected checkpoint. }
% \label{app_dev_corr}
% \vspace{-0.4cm}
% \end{figure*}

% \begin{figure*}[ht]
% \centering
% \subfigure[Analysis on Qwen2.5$_{7B}$.]
% {\includegraphics[scale=0.7]{Figure/dev_cons_qw2.5.pdf}
% }
% \subfigure[Analysis on DeepSeek-R1$_{7B}$.]
% {\includegraphics[scale=0.7]{Figure/dev_cons_dsqw.pdf}
% }
% \subfigure[Analysis on LLaMA3$_{8B}$.]
% {\includegraphics[scale=0.7]{Figure/dev_cons_ll3.pdf}
% }
% \subfigure[Analysis on Qwen3$_{4B}$.]
% {\includegraphics[scale=0.7]{Figure/dev_cons_qw3.pdf}
% }
% \caption{For each iteration, performance on the validation set under our S$^{3}$-CoT$^{sc}$ method. Green: Accuracy; Yellow: Length; Red Circle: Our selected checkpoint.}
% \label{app_dev_cons}
% \vspace{-0.4cm}
% \end{figure*}

\end{document}